\documentclass{article}

\PassOptionsToPackage{numbers, compress,sort}{natbib}
\usepackage[preprint]{neurips_2025}
\usepackage{natbib}
\usepackage{pdfpages}
\usepackage{adjustbox}
\usepackage{multirow}
\usepackage{tikz}
\usepackage{makecell}
\usepackage{amsmath}
\usepackage{bm}
\usepackage{hyperref}
\usepackage{cleveref}
\usepackage{cancel}
\usepackage{wrapfig}
\usepackage{xcolor}

\usepackage[utf8]{inputenc} 
\usepackage[T1]{fontenc}    
\usepackage{url}            
\usepackage{booktabs}       
\usepackage{amsfonts}       
\usepackage{nicefrac}       
\usepackage{microtype}      
\usepackage{xcolor}         
\usepackage[ruled,vlined]{algorithm2e}

\usepackage{mathtools}

\DeclareMathOperator*{\argmin}{arg\,min}

\title{FAROS: Fair Graph Generation via Attribute Switching Mechanisms}

\newcommand{\circlemarker}[1]{\tikz\draw[fill=#1, draw=#1] (0,0) circle (0.8ex);}

\newcommand{\squaremarker}[1]{\tikz\draw[fill=#1, draw=#1] (0,0) rectangle (0.24cm,0.24cm);}

\newcommand{\trianglemarker}[1]{%
  \tikz\draw[fill=#1, draw=#1] 
    (0,0) -- (0.24cm,0) -- (0.12cm,0.24cm) -- cycle;}

\definecolor{forest}{RGB}{34,139,34}

\author{Abdennacer Badaoui\thanks{Equal contribution.} \quad Oussama Kharouiche\footnotemark[1] \quad Hatim Mrabet\footnotemark[1] \\  \textbf{Daniele Malitesta}\footnotemark[1] \quad \textbf{Fragkiskos D. Malliaros} \\ ~\\
Université Paris-Saclay, CentraleSupélec, Inria, France \\
\texttt{\{abdennacer.badaoui, oussama.kharouiche, hatim.mrabet\}@student-cs.fr}\\
\texttt{\{daniele.malitesta, fragkiskos.malliaros\}@centralesupelec.fr
}}

\begin{document}

\maketitle

\begin{abstract} 
Recent advancements in graph diffusion models (GDMs) have enabled the synthesis of realistic network structures, yet ensuring fairness in the generated data remains a critical challenge. Existing solutions attempt to mitigate bias by re-training the GDMs with ad-hoc fairness constraints. Conversely, with this work, we propose \textbf{FAROS}, a novel \underline{\textbf{FA}}ir graph gene\underline{\textbf{R}}ati\underline{\textbf{O}}n framework leveraging attribute \underline{\textbf{S}}witching mechanisms and directly running in the generation process of the pre-trained GDM. Technically, our approach works by altering nodes' sensitive attributes during the generation. To this end, FAROS calculates the optimal fraction of switching nodes, and selects the diffusion step to perform the switch by setting tailored multi-criteria constraints to preserve the node-topology profile from the original distribution (a proxy for accuracy) while ensuring the edge independence on the sensitive attributes for the generated graph (a proxy for fairness). Our experiments on benchmark datasets for link prediction demonstrate that the proposed approach effectively reduces fairness discrepancies while maintaining comparable (or even higher) accuracy performance to other similar baselines. Noteworthy, FAROS is also able to strike a better accuracy-fairness trade-off than other competitors in some of the tested settings under the Pareto optimality concept, demonstrating the effectiveness of the imposed multi-criteria constraints.
\end{abstract}

\section{Introduction}

As much of the data in machine learning is represented through graph structures, graph machine learning has found widespread application in several fields~\cite{DBLP:series/synthesis/2020Hamilton}, such as recommender systems~\cite{DBLP:conf/sigir/0001DWLZ020, DBLP:conf/iclr/Cai0XR23}, drug-target interaction prediction~\cite{DBLP:journals/tcbb/EzzatZW0K17, DBLP:journals/artmed/NingWZSJWY25}, or spatiotemporal forecasting~\cite{gegenGNN-TNNLS24}. By exploiting the graph topology, powerful representation learning models such as graph neural networks (GNNs)~\cite{DBLP:journals/tnn/ScarselliGTHM09} have settled as the de facto approaches in machine learning, largely outperforming previous prediction techniques (e.g., multi-layer perceptrons) unable to see data as nodes interconnected through edges~\cite{DBLP:conf/iclr/YangWWY23}.

\textbf{Fairness in graph data.} Despite the new graph machine learning wave, graph-based models (like any other traditional machine learning models~\cite{DBLP:journals/csur/CatonH24}) struggle when dealing with bias intrinsic within the data, leading to unfair prediction outcomes~\cite{DBLP:conf/wsdm/DaiW21, DBLP:conf/iclr/LiWZHL21, DBLP:journals/tai/SpinelliSHU22, DBLP:journals/tkdd/KoseS24, DBLP:conf/aaai/WangCDWWPZ25}. In fact, the issue is even amplified by the topological graph structure. A trivial but concrete example lies in the job recommendation domain, where a graph-based model would suggest the same job to people of the same gender by following the homophily pattern: people of the same gender tend to be closely connected within the graph topology~\cite{DBLP:conf/cikm/ZehlikeB0HMB17}. Indeed, there is extensive literature in graph machine learning regarding fairness, how to measure it, and how to mitigate it~\cite{DBLP:journals/tkdd/ChenRPTWYKDA24, DBLP:journals/corr/abs-2205-05396, DBLP:journals/tkde/DongMWCL23}. In general, and differently from other machine learning fields and tasks, fairness in graph data is defined on two levels. First, it may depend on sensitive information contained in node features. Second, it is also influenced by the graph topology---how nodes are connected across multiple hops.

\textbf{Graph diffusion models.} Diffusion models (DMs)~\cite{DBLP:journals/tkde/CaoTGXCHL24, DBLP:journals/csur/XingFCDHXWJ25} have gained significant traction in generative machine learning, following earlier successes with VAEs~\cite{DBLP:journals/corr/KingmaW13} and GANs~\cite{DBLP:journals/corr/GoodfellowPMXWOCB14}. Initially successful in computer vision for tasks like image generation~\cite{DBLP:journals/air/ChenXHYYCZ25, DBLP:conf/icml/RameshPGGVRCS21, DBLP:conf/cvpr/RombachBLEO22}, semantic segmentation~\cite{DBLP:conf/iclr/Child21}, and image super-resolution~\cite{DBLP:journals/pami/SahariaHCSFN23}, DMs have also been applied to natural language~\cite{DBLP:conf/acl/HanKT23, DBLP:conf/nips/LiTGLH22} and multi-modal generation~\cite{DBLP:journals/corr/abs-2204-06125, DBLP:conf/icml/0006YMXE024}. The success of DMs has extended to graph machine learning~\cite{DBLP:conf/ijcai/LiuFLLLLTL23, DBLP:conf/log/0001DWXZLW22, DBLP:journals/pami/GuoZ23}, where graph diffusion models (GDMs) are used to generate new graph structures, particularly in drug discovery and protein design~\cite{DBLP:conf/icml/HoogeboomSVW22, DBLP:conf/iclr/XuY0SE022, DBLP:journals/corr/abs-2304-01565} and to improve graph machine learning capabilities~\cite{DBLP:conf/nips/YangYZS23}. GDMs adapt the diffusion process for graph data, operating in either continuous~\cite{DBLP:conf/icml/JoLH22, DBLP:conf/aistats/NiuSSZGE20, DBLP:conf/icdm/HuangS0FL22} or discrete spaces~\cite{DBLP:conf/iclr/VignacKSWCF23, DBLP:conf/nips/0007QCCFP0DT24, DBLP:journals/corr/abs-2401-13858,siraudin2025cometh}. Early GDMs were primarily limited to generating small graphs. More recent methods address the generation of large attributed graphs by employing mini-batching for edge generation (GraphMaker~\cite{DBLP:journals/corr/abs-2310-13833}) and performing diffusion on sampled subgraphs~\cite{DBLP:conf/icml/TrivediRA0DKLPA24}.

\textbf{Fairness in (graph) diffusion models.} Several studies have identified and begun to address bias in visual diffusion models, noting that real-world dataset bias can be amplified during generation. For example, Stable Diffusion has been shown to produce low-diversity images \cite{DBLP:journals/corr/abs-2302-10893}, and prompts with trait annotations often yield stereotyped or skewed portrayals in terms of skin tone and occupation \cite{DBLP:conf/fat/0001KDLCNHJ0C23,DBLP:conf/iccv/0001ZB23}. Similar fairness issues have also been recognized in the recommendation scenario \cite{DBLP:journals/corr/abs-2409-04339}. Proposed mitigations include fair-guidance instructions \cite{DBLP:journals/corr/abs-2302-10893}, dynamic attribute switching \cite{DBLP:conf/aaai/Choi0K0P24}, ad-hoc fine-tuning \cite{DBLP:journals/corr/abs-2311-07604}, and adaptive latent guidance \cite{DBLP:journals/corr/abs-2503-01872}. In contrast, fairness in graph diffusion models remains underexplored. Recent efforts estimate unbiased edges from node attributes \cite{DBLP:conf/eusipco/NavarroRBMS24,DBLP:conf/nips/NavarroRBMS24}, while theoretical analyses examine bias sources in graph generation; notably, FairGen \cite{DBLP:conf/icde/Zheng0TXZH24} targets non-diffusion generators, and FairWire \cite{DBLP:conf/nips/KoseS24} evaluates downstream link prediction performance.

\textbf{Motivation.} While existing fairness-aware methods achieve promising results by re-training the GDMs to enforce fairness, they often overlook the temporal dynamics of the generation process, failing to identify and intervene at the critical timestep in the pipeline to mitigate unfairness. This might excessively limit the expressiveness of the GDM they are bind to, leading to suboptimal trade-offs in terms of accuracy-fairness in the downstream task. Conversely, we maintain that targeted interventions to the generation process, without any re-training of the GDM, could potentially lead to more balanced performance trends in the downstream task (as outlined in our results).

\begin{wrapfigure}{R}{0.5\textwidth}
  \centering
\includegraphics[width=0.4\textwidth]{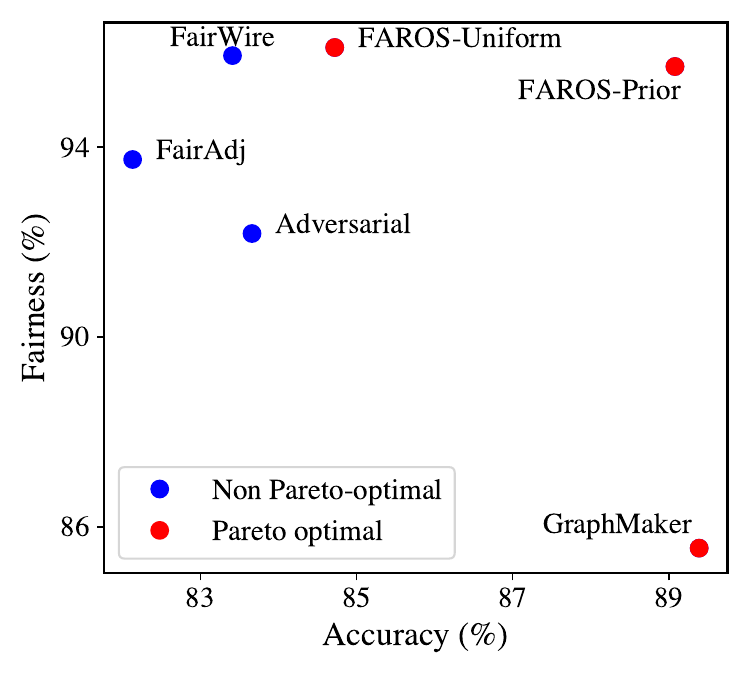} 
    \caption{Link prediction results in terms of accuracy (AUC) and fairness (100 - $\Delta_{EO}$) for \textsc{Cora} with the tested approaches trained on the generated graph data. In the plot, (\textcolor{blue}{blue}) \textcolor{red}{red} points are solutions (not) belonging to the Pareto frontier. FAROS clearly strikes a better accuracy-fairness trade-off than other baselines.}
    \label{fig:pareto_introduction}
\end{wrapfigure}

\textbf{Our work.} To this end, we propose a novel \underline{\textbf{FA}}irness graph gene\underline{\textbf{R}}ati\underline{\textbf{O}}n framework leveraging attribute \underline{\textbf{S}}witching mechanisms (\textbf{FAROS}). The model alters sensitive attributes of graph nodes during the GDM generation process. Importantly, we aim to enhance fairness with minimal modifications to the generation process, requiring no re-training of the pre-trained GDM. Rather than switching attributes of all nodes, FAROS mathematically calculates the optimal fraction of nodes for attribute switching by analyzing differences between newly-generated links connecting same- and different-group nodes. We then select the optimal time step for attribute switching, optimizing a multi-criteria function that preserves the original node-topology profile while ensuring edge independence from sensitive attributes. Benchmarking results for link prediction demonstrate FAROS's ability to mitigate fairness issues while maintaining comparable or improved accuracy. Notably, our approach achieves better accuracy-fairness trade-offs, in some cases even under Pareto optimality criteria  (see \Cref{fig:pareto_introduction} for highlighted results and the Supplementary Material for all Pareto plots), further validating our multi-criteria constraints.
\section{Preliminaries}

In this section, we provide the useful notions as necessary background for our proposed methodology, FAROS. First, we introduce some common notations and definitions for fairness in link prediction tasks~\cite{DBLP:journals/tkdd/ChenRPTWYKDA24, DBLP:journals/corr/abs-2205-05396, DBLP:journals/tkde/DongMWCL23}. Then, we describe graph diffusion models (GDMs), with a particular focus on GraphMaker~\cite{DBLP:journals/corr/abs-2310-13833}, which we adopt in our experiments as it currently represents the state-of-the-art in large graph generation with diffusion models~\cite{DBLP:conf/nips/KoseS24}. Finally, we present selected approaches from the literature which propose to address fairness in DMs~\cite{DBLP:conf/aaai/Choi0K0P24} and GDMs~\cite{DBLP:conf/nips/KoseS24}.

\textbf{Notation.}  Let $\mathcal{G} = \{\mathcal{V}, \mathcal{E}\}$ be an undirected graph with node set $\mathcal{V} = \{v_1, v_2, \ldots, v_N\}$ and edge set $\mathcal{E} \subseteq \mathcal{V} \times \mathcal{V}$. Then, we define the graph adjacency matrix $\mathbf{A} \in \{0, 1\}^{N \times N}$, where $A_{ij} = 1$ if and only if $(v_i, v_j) \in \mathcal{E}$, 0 otherwise. Moreover, we introduce $\mathbf{X} \in \mathbb{R}^{N \times F}$ as the node features, where $\mathbf{X}_i \in \mathbb{R}^{F}$ is the feature vector of node $v_i$, and $S_i$ as the sensitive attribute of node $v_i$ taking a value from the set $\mathcal{S} = \{s_1, s_2, \ldots, s_K\}$ to represent one of $K$ categories in a multi-class scenario (e.g., different demographic groups or affiliations). Finally, for graph diffusion models, we use the superscript to indicate the specific time step during the forward/reverse processes or generation (e.g., $\mathcal{G}^{(t-1)}$ referring to perturbed/denoised graph at time $t-1$). Conversely, the final generated data, as well as the predicted values, are denoted with a tilde, such as $\tilde{\mathbf{A}}$ or $\tilde{\mathcal{G}}$.

\textbf{Fairness in link prediction.} We provide the definitions of two popular fairness metrics, statistical parity and equality of opportunity, in the task of link prediction~\cite{DBLP:journals/tkdd/ChenRPTWYKDA24, DBLP:journals/corr/abs-2205-05396, DBLP:journals/tkde/DongMWCL23} (the considered task for this work). Let $\tilde A_{ij}\in\{0,1\}$ be the predicted link between nodes $v_i,v_j$, and let $S_i,S_j$ be their sensitive attributes. We denote by $\mathcal U$ the uniform distribution over $\mathcal V\times\mathcal V$.

\quad 1. \textit{Statistical parity} ($\Delta_{SP}$): it measures the absolute difference in predicted link rates for same-group vs. different-group pairs; a smaller $\Delta_{SP}$ means the predictor does not favor one pairing type over the other:
\begin{equation}
\Delta_{SP}
= \biggl|\,
\mathbb{E}_{(v_i,v_j)\sim\mathcal U}\bigl[\tilde A_{ij}=1 \mid S_i=S_j\bigr]
\;-\;
\mathbb{E}_{(v_i,v_j)\sim\mathcal U}\bigl[\tilde A_{ij}=1 \mid S_i\neq S_j\bigr]
\biggr|.
\end{equation}
\quad 2. \textit{Equality of opportunity} ($\Delta_{EO}$): it measures the absolute difference in true-positive rates (only among ground-truth links) for same-group vs. different-group pairs; a lower $\Delta_{EO}$ indicates fairer correct-prediction rates across groups:
\begin{equation}
\Delta_{EO}
= \biggl|\,
\mathbb{E}_{(v_i, v_j)\sim\mathcal U}\bigl[\tilde A_{ij}=1 \mid A_{ij}=1,\,S_i=S_j\bigr]
\;-\;
\mathbb{E}_{(v_i, v_j)\sim\mathcal U}\bigl[\tilde A_{ij}=1 \mid A_{ij}=1,\,S_i\neq S_j\bigr]
\biggr|.
\end{equation}



\textbf{Graph diffusion models for large attributed graph generation.} GraphMaker~\cite{DBLP:journals/corr/abs-2310-13833} currently constitutes the state-of-the-art GDM to generate large attributed graphs. Let $\mathbf{X} \in \mathbb{R}^{N \times F \times C}$ and $\mathbf{A} \in \mathbb{R}^{N \times N \times 2}$ be the one-hot encoded representations of the node features and the adjacency matrix, respectively, where $C$ is the number of possible classes an attribute can have.
 During the forward process, the model corrupts both the adjacency matrix and the node features such as, for $t = 1, 2, \ldots, T$, for any $ i \in [N],\ f \in [F]$ we have $q(\mathbf{X}_{if}^{(t)} \; | \; \mathbf{X}^{(0)}_{if}) = \mathbf{X}^{(0)}_{if}\overline{\mathbf{Q}}_{X_f}^{(t)}$ and $q(\mathbf{A}^{(t)} \; | \; \mathbf{A}^{(0)}) = \mathbf{A}^{(0)}\overline{\mathbf{Q}}_{A}^{(t)}$, where $\overline{\mathbf{Q}}_{X_f}^{(t)}$ and $\overline{\mathbf{Q}}_{A}^{(t)}$ are the forward process transition matrices. Then, in the reverse process, a denoising network $\theta$ is trained to reconstruct the original graph data from the noise distribution through $p_{\theta}(\mathcal{G}^{(t - 1)} \; | \; \mathcal{G}^{(t)}, t) = \prod_{i = 1}^{N} \prod_{f = 1}^{F} p_{\theta}(\mathbf{X}_{if}^{(t-1)} \; | \; \mathcal{G}^{(t)}, t) \prod_{1 \leq i \leq j \leq N} p_{\theta}(\mathbf{A}_{ij}^{(t - 1)} \; | \; \mathcal{G}^{(t)}, t)$, where $\mathcal{G}^{(t)}$ is the whole graph for $t = T, T-1, \ldots, 1$. Two variant schedules are considered: synchronous (\textbf{Sync}), which jointly restores $\mathbf{X}$ and $\mathbf{A}$ across every denoising step; and asynchronous (\textbf{Async}), which decouples the process by recovering $\mathbf{X}$ first, and then $\mathbf{A}$, to enhance their mutual correlation.


\textbf{Fair sampling via attribute switching and FairWire.} The approach in~\cite{DBLP:conf/aaai/Choi0K0P24} proposes to modify the generation process of the DM for image generation by selecting an optimal time step $t = \tau^{*}$ when to switch the sensitive attribute from the original to a new value $s_{\text{org}} \xrightarrow{\text{switch}} s_{\text{new}}$, with $s_{\text{org}}, s_{\text{new}} \in \mathcal{S} = \{s_1, s_2\}$ as in a binary setup. 
Thus, the attribute switching framework identifies the time step $\tau^{*}$ to switch the sensitive attribute.  It adopts $p_{\theta}(\mathbf{X}_i^{(t-1)} \; | \; \mathbf{X}_i^{(t)}, S_{i} = s_{\text{org}})$ for $t = T, T-1, \ldots, \tau^* + 1$, and then it switches to $p_{\theta}(\mathbf{X}_i^{(t-1)} \; | \; \mathbf{X}_i^{(t)}, S_i = s_{\text{new}})$ for $t = \tau^*, \tau^* -1, \ldots, 1$, where $\mathbf{X}^{(t)}_i$ is an image in this case. Noteworthy, the DM is pre-trained, so the operation runs exclusively during the generation process. In our framework, FAROS, we switch nodes' sensitive attributes during the generation process of a pre-trained GDM. However, unlike the approach in~\cite{DBLP:conf/aaai/Choi0K0P24}, we perform the switching on an optimal fraction of nodes from the generated graph, select the optimal time step to perform the switching through a multi-criteria optimization function, and switch the sensitive attributes with more ad-hoc strategies by sampling from multinomial distributions. 

FairWire~\cite{DBLP:conf/nips/KoseS24} extends GDMs by introducing a fairness regularizer which is coupled with the main loss function of the GDM to penalize deviations from a balance condition linked to statistical parity. Specifically, the regularizer controls the ratio between intra-group and inter-group edges during generation. Unlike FairWire, our approach, FAROS, does not require re-training the underlying GDM. Instead, it leverages attribute switching mechanisms to modify sensitive node attributes during the generation process of the pre-trained GDM.

\section{Fair Graph Generation with FAROS}

We present FAROS, a framework for fairness-aware generation that leverages attribute switching mechanisms~\cite{DBLP:conf/aaai/Choi0K0P24}. With FAROS, our purpose is to drive the generation of a GDM towards fairness without drastically changing its functioning. That is why FAROS acts during the generation phase of an already pre-trained GDM (unlike FairWire~\cite{DBLP:conf/nips/KoseS24}) and dynamically alters the sensitive attributes of a limited optimal fraction of nodes from the generated graph. In the following, we first detail how we estimate the optimal fraction $\rho^*$ of nodes to switch, balancing the expected bias amplification effect introduced by the GDM. Later, we present our tailored attribute switching mechanisms on generated graph nodes. Finally, we define a multi-criteria optimization problem to select the optimal time step $\tau^*$ to perform the switch, which works by balancing the preservation of the original node-structure profile (a proxy for accuracy) against edge–attribute independence (a proxy for fairness). 

\subsection{Estimation of the optimal fraction $\rho^{*}$}
\label{sec:rho_estimation}

As discussed above, our main purpose is to drive the GDM generation towards fairness by affecting its functioning the least. Therefore, we choose to perform attribute switching on a subset of nodes $\mathcal{V}^* \subseteq \mathcal{V}$ from the generated graph, rather than on the whole node set $\mathcal{V}$. This involves estimating the optimal fraction of these nodes, $\rho^* = \mathbb{E}\left[|\mathcal{V}^*|/|\mathcal{V}|\right]$, for attribute switching, in order to counteract the bias amplification effect introduced by the GDM during the generation process. 

In our intuition, this amplification is caused by an imbalance between the generated exterior links (those connecting nodes with different sensitive attributes) and interior links (those connecting nodes with the same sensitive attributes). Thus, we aim to find the optimal fraction of nodes to be switched such that the generated imbalance goes in the opposite direction of its natural tendency, trying to mitigate its harmful effects. This involves calculating the number of interior and exterior links before generation, and estimating the expected number of interior and exterior links after the generation.

Let $\mathcal{E}_{\text{int}} = \{e_{ij} \in \mathcal{E} \; | \; S_i = S_j \}$ and $\mathcal{E}_{\text{ext}} = \{e_{ij} \in \mathcal{E} \; | \; S_i \neq S_j \}$ be the sets of interior and exterior links before the generation; dually, $\tilde{\mathcal{E}}_{\text{int}} = \{\tilde{e}_{ij} \in \tilde{\mathcal{E}} \; | \; \tilde{S}_i = \tilde{S}_j \}$ and $\tilde{\mathcal{E}}_{\text{ext}} = \{\tilde{e}_{ij} \in \tilde{\mathcal{E}} \; | \; \tilde{S}_i \neq \tilde{S}_j \}$ are the interior and exterior links after the generation. For a fixed fraction $\rho \in [0,1]$, we obtain the following expectations as quadratic polynomials for $\rho$: $\mathbb{E}[|\tilde{\mathcal{E}}_{\text{ext}}|] = R^{2}_{\text{ext}}\rho^{2} + R^{1}_{\text{ext}}\rho + |\mathcal{E}_{\text{ext}}|$ and $\mathbb{E}[|\tilde{\mathcal{E}}_{\text{int}}|] = R^{2}_{\text{int}}\rho^{2} + R^{1}_{\text{int}}\rho + |\mathcal{E}_{\text{int}}|$, where $R^2_{\text{int}}, R^1_{\text{int}}$ and $R^2_{\text{ext}}, R^1_{\text{ext}}$ represent the coefficients of the two polynomials obtained through some re-working. Finally, to counteract the above-described imbalance, we aim to solve the following quadratic optimization problem for $\rho$:
\begin{equation}
\label{eq:opt_frac}
\rho^{*} = \argmin_{\rho \in [0, 1]} \text{sign}(R_0)(R_2\rho^{2} + R_1\rho + R_0),
\end{equation}
where $R_2 =  R_{\text{ext}}^{2} - R_{\text{int}}^{2}$, $R_1 = R_{\text{ext}}^{1} - R_{\text{int}}^{1}$, $R_0 = |\mathcal{E}_{\text{ext}}| - |\mathcal{E}_{\text{int}}|$, while $\text{sign}(R_0)$ stands for the sign of $R_0$ and serves to drive the optimization in the opposite direction of the imbalance. Further details on how this formulation was derived can be found in the Supplementary Material.

\subsection{Tailored attribute switching on generated graph nodes}


At this point, we perform the graph generation process by applying the attribute switching mechanisms on the optimal fraction of nodes $\rho^*$. The generation runs without any modification (pre-switching generation) until a specific time step $t = \tau^*$, and then it involves the attribute switching mechanisms (post-switching generation). In the following, we provide all the details regarding our methodology.  

\textbf{Selection of switching nodes.} We aim to select a random subset of nodes $\mathcal{V}^* \subseteq \mathcal{V}$ to perform the attribute switching, based on the estimated fraction $\rho^*$ (\Cref{eq:opt_frac}). To this end, we sample $\mathcal{V}^* = \{v \in \mathcal{V} \mid z_v = 1\}$ \text{with} $z_v \sim \mathcal{B}(\rho^*)$ i.i.d., where $\mathcal{B}$ is a Bernoulli distribution to ensure that node sampling occurs independently.


\textbf{Sampling of the new sensitive attribute.} For each node $v_i$ in the selected set $\mathcal{V}^*$, we sample a new value $s_{\text{new}} \leftarrow s_{\text{samp}} \sim \mathcal{S} \setminus \{s_{\text{org}}\}$ via a distribution $\mathcal{D}_{\sigma}$, where we avoid sampling the original value ($s_{\text{org}}$) to ensure the effectiveness of the attribute switching. We consider two choices for $\mathcal{D}_{\sigma}$: in uniform sampling, we have $\mathcal{U}(S = s_{\text{new}} \mid s_{\text{new}} \neq s_{\text{org}}) = \frac{1}{|\mathcal{S}| - 1}$, while in prior sampling, we have $\mathcal{P}(S = s_{\text{new}} \mid s_{\text{new}} \neq s_{\text{org}}) = \frac{p_{\theta}(S = s_{\text{new}})}{\sum_{s_{\text{new}}' \in \mathcal{S}, s_{\text{new}}' \neq s_{\text{org}}} p_\theta(S = s_{\text{new}}')}$. Conversely, non-switched nodes retain their original attribute ($s_{\text{new}} \leftarrow s_{\text{org}}$). 


\textbf{Nodes attribute switching mechanisms.} We use the pre-trained GDM to initiate the generation of the graph data. Thus, the usual generation procedure is followed, obtaining $\mathcal{G}^{(t-1)}$ through $p_\theta(\mathcal{G}^{(t-1)} \; | \; \mathcal{G}^{(t)}, S = s_{\text{org}})$ for $t = T, T -1, \ldots, \tau^* + 1$. Then, at $t = \tau^*$, FAROS samples the subset of optimal nodes $\mathcal{V}^*$, as well as the new sensitive attribute for each selected node according to the chosen sampling distribution $\mathcal{D}_{\sigma}$. From this time step on, so for $t = \tau^*, \tau^* - 1, \ldots, 1$, FAROS keeps generating the graph data through $p_\theta(\mathcal{G}^{(t-1)} \; | \; \mathcal{G}^{(t)}, S = s_{\text{new}})$. In the following, we will indicate the generated graph with optimal node fraction $\rho^*$ and optimal switching time $\tau^*$ as $\tilde{\mathcal{G}}_{\rho^*, \tau^*}$.



\subsection{Multi-criteria selection of the optimal sampling step $\tau^{*}$}
\label{sec:multi-criteria}

To complete the overall pipeline presentation, we detail the selection process of the switching time step $\tau$ to find the optimal value $\tau^*$. To this end, we designed a multi-criteria optimization method based upon two objectives, namely, the preservation of the original node-structure profile (a proxy for accuracy) and the edge independence on the sensitive attributes for the generated graph (a proxy for fairness). We remind that both accuracy and fairness are considered with relation to the final downstream task we are pursuing, namely, link prediction.

\textbf{Node-topology preservation (accuracy).} For the first objective, we seek to generate new graphs whose probability distribution does not deviate from the original one learned by the GDM. We assume this represents a good proxy to ensure high accuracy performance in the downstream task where the generated graph will constitute the training set. Since the considered GDM~\cite{DBLP:journals/corr/abs-2310-13833} is generating both topology and node features, we aim to choose a distance measure between the original and generated graph distributions which can capture both aspects simultaneously.

We choose the Fused Gromov-Wasserstein (FGW) distance~\cite{DBLP:conf/icml/VayerCTCF19}, adopted to compare graphs through their node features and structural information by combining the Wasserstein distance (accounting for node features only) and the Gromov-Wasserstein distance (accounting for topology only). Concretely, the idea is to calculate the FGW distance by solving the optimal transport problem to shift objects from one location to another (in our case, the original graph $\mathcal{G}$ and any generated graph $\tilde{\mathcal{G}}$). In this formulation, the two graphs are represented through their associated probability distributions, which we denote, by a slight abuse of notation, as $\mathcal{G}$ and $\tilde{\mathcal{G}}$. Let $\mathbf{M} \in \mathbb{R}^{N \times N}$ be the transportation cost matrix for node features in the Wasserstein distance, and let $\mathbf{C}_{\mathcal{G}}, \mathbf{C}_{\tilde{\mathcal{G}}} \in \mathbb{R}^{N \times N}$ be the internal cost matrices for the original and generated graphs $\mathcal{G}$ and $\tilde{\mathcal{G}}$ as in the Gromov-Wasserstein distance. Through a coefficient $\alpha \in [0, 1]$ to set the importance of the two measures, we calculate the FGW distance as: $\text{FGW}(\mathcal{G}, \tilde{\mathcal{G}}) = \min_{\mathbf{P} \in \prod(\mathcal{G}, \tilde{\mathcal{G}})} \langle \alpha\mathbf{M}+(1-\alpha)|\mathbf{C}_{\mathcal{G}}(i, k)-\mathbf{C}_{\tilde{\mathcal{G}}}(j, l)|^2 \otimes \mathbf{P}, \mathbf{P} \rangle$, where $\mathbf{P} \in \mathbb{R}^{N \times N}$ is a joint probability distribution of the two graphs. Further details on how we adapted the original formulation of the FGW to our specific case are provided in the Supplementary Material.

\textbf{Edge independence (fairness).} For the second objective, we aim to generate new graphs whose edges are independent on the sensitive attributes of the endpoint nodes. We assume this represents a good proxy to ensure high fairness performance in the downstream task, where the generated graph will constitute the training set. Let $(v_i, v_j)$, $i \neq j$ be a pair of nodes in any generated graph, and $S_i, S_j$ their respective sensitive attributes. Ideally, for a persfect edge independence on the sensitive attributes, we seek to have, for any new created link in the generated graph and $s \in \mathcal{S}$: $p(S_j = s \; | \; S_i = s, \tilde{A}_{ij} = 1) = p(S_j \neq s \; | \; S_i = s, \tilde{A}_{ij} = 1) = 1/2$. 

To this end, we define the entropy as a measure to encourage this uniform distribution, where the entropy is maximized when the probabilities are equal to $1/2$, indicating maximal edge independence for the sensitive attribute value $s \in \mathcal{S}$. We calculate the entropy over all the sensitive attributes in the generated graph $\tilde{\mathcal{G}}$ as $\text{H}(\tilde{\mathcal{G}}) = \sum_{s \in \mathcal{S}} [-p_{ss} \text{log}(p_{ss}) - p_{s\overline{s}}\text{log}(p_{s\overline{s}}))],$ where we define, for a fixed  $s \in \mathcal{S}$,  $p_{ss} = p(S_j = s \; | \; S_i = s, \tilde{A}_{ij} = 1)$ and $p_{s\overline{s}} = p(S_j \neq s \; | \; S_i = s, \tilde{A}_{ij} = 1)$.

\textbf{Multi-criteria optimization problem.} Let us denote the set of admissible values for the optimal switching time step as $\{T-1, \ldots, 1\}$, where we excluded $\tau = T$ because no generation under the prior distribution has occurred at this point, making a switch meaningless. For a given optimal fraction $\rho^{*}$ calculated by \Cref{eq:opt_frac}, and leveraging the definitions from above, we introduce the overall multi-criteria optimization problem:
\begin{equation}
    \tau^* = \argmin_{\tau \in \{T-1, \ldots, 1\}}  \underbrace{\text{FGW}(\mathcal{G}, \tilde{\mathcal{G}}_{\rho^*, \tau})}_{\text{\textbf{accuracy}}} - \gamma \underbrace{\text{H}(\tilde{\mathcal{G}}_{\rho^*, \tau})}_{\text{\textbf{fairness}}}.
    \label{eq:opt_tau}
\end{equation}
The parameter $\gamma \in [0, 1]$ balances the trade-off between accuracy, represented by the FGW term (node-topology distance), and fairness, represented by the entropy term H; higher values of $\gamma$ give more importance to fairness.

\begin{figure}[!t]
    \centering    \includegraphics[width=\columnwidth]{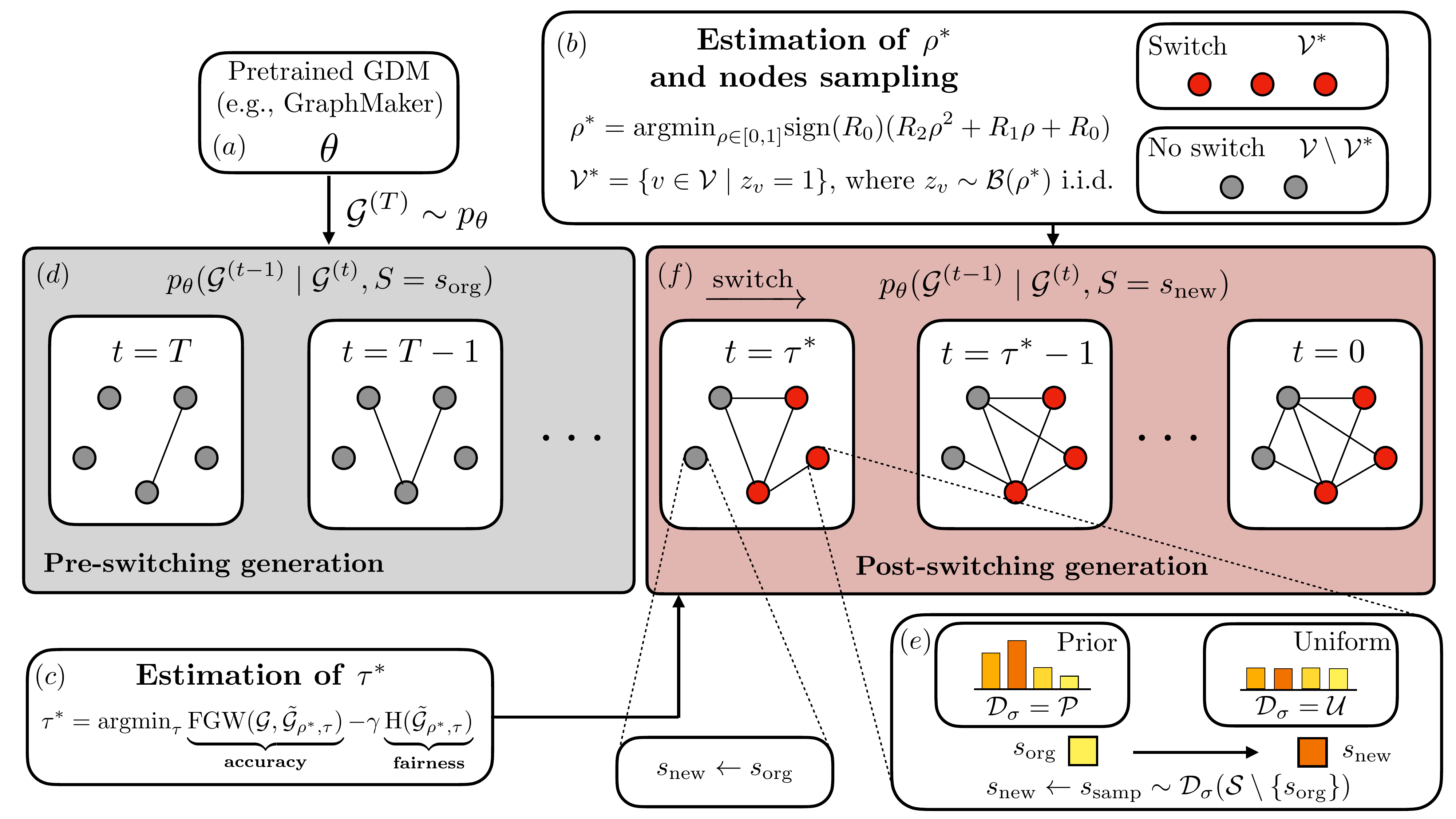} 
    \caption{Overall pipeline of FAROS. By starting from a pre-trained GDM $(a)$, we first estimate the optimal fraction of nodes to switch and sample them $(b)$. Second, we estimate the optimal time step at which to perform the switching $(c)$. Then, the generation process starts without any modification $(d)$ until $t = \tau^*$, when we perform the attribute switching $(e)$. From that moment until the end, the generation continues without any other switching $(f)$.}
    \label{fig:model}
\end{figure}

To conclude, we provide an illustration of the overall pipeline of FAROS (\Cref{fig:model}), along with its pseudocode (Algorithm \ref{alg:method}).

\let\oldemptyset\emptyset

\begin{algorithm}[!t]
\caption{FAROS fair graph generation algorithm.}
\label{alg:method}
\SetAlgoLined
\textbf{Input:} Pre-trained GDM $\theta$, prior noise distribution $p_{\theta}$, selected attribute switching distribution $\mathcal{D}_{\sigma} = \{\mathcal{U}, \mathcal{P}\}$, admissible set of values for the sensitive attribute $\mathcal{S}$. \\
\textbf{Output:} Generated fair graph $\tilde{\mathcal{G}}$. \\
Estimate $\rho^{*}$ through \Cref{eq:opt_frac}\\
Estimate $\tau^{*}$ through \Cref{eq:opt_tau}\\
Initialize $\mathcal{G}^{(T)} \sim p_{\theta}$\\
\For{$t = T, T - 1, \ldots, \tau^{*}+1$}{
Get $\mathcal{G}^{(t-1)}$ through $p_\theta(\mathcal{G}^{(t-1)} \; | \; \mathcal{G}^{(t)}, S = s_{\text{org}})$ \phantomsection{    } \textbf{\# Pre-switching generation}\\
}
Sample $\mathcal{V}^* = \{v \in \mathcal{V} \mid z_v = 1\}$ \text{where} $z_v \sim \mathcal{B}(\rho^*)$ i.i.d.\\
\For{$v_i \in \mathcal{V}$}{
Sample $s_{\text{samp}} \sim \mathcal{D}_\sigma(\mathcal{S} \setminus \{s_{\text{org}}\})$ \\
Update $s_{\text{new}} \leftarrow s_{\text{samp}} \text{ if } v_{i} \in \mathcal{V}^* \text{ else } s_{\text{new}} \leftarrow s_{\text{org}}$ \phantomsection{    }\textbf{\# Attribute switching} \\
}
\For{$t = \tau^*, \tau^* - 1, \ldots, 1$}{
Get $\mathcal{G}^{(t-1)}$ through $p_\theta(\mathcal{G}^{(t-1)} \; | \; \mathcal{G}^{(t)}, S = s_{\text{new}})$ \phantomsection{    } \textbf{\# Post-switching generation}\\
}
\textbf{Return:} $\tilde{\mathcal{G}}$.
\end{algorithm}

\section{Experiments and Results}
\label{sec:results}

\subsection{Experimental setup}

\textbf{Datasets.} For fair comparison to our main baseline, FairWire~\cite{DBLP:conf/nips/KoseS24}, we use the same datasets for graph generation and link prediction as in the original paper, namely, \textsc{Cora}, \textsc{Citeseer}, and \textsc{Amazon Photo}. \textsc{Cora} and \textsc{Citeseer} are standard datasets for link prediction, representing two citation networks where nodes represent research papers and the edges the citations between them~\cite{DBLP:journals/aim/SenNBGGE08}. Conversely, \textsc{Amazon Photo}~\cite{DBLP:journals/corr/abs-1811-05868} refers to a product category of the Amazon reviews dataset where items bought together are linked within the graph. 

\textbf{Baselines.} We decide to exploit the same baselines as in FairWire~\cite{DBLP:conf/nips/KoseS24} for the task of graph generation, as they currently represent the state-of-the-art approaches in fairness-aware graph machine learning and fair graph generation. First, we use GraphMaker~\cite{DBLP:journals/corr/abs-2310-13833} (in the synchronous version) as the fairness-agnostic graph generator and link predictor. Then, we use FairAdj~\cite{DBLP:conf/iclr/LiWZHL21} and an adversarial regularizer~\cite{DBLP:conf/wsdm/DaiW21} (referred to as Adversarial) representing a post-processing and an in-training fairness-aware link prediction model, respectively. In the proposed model categorization, FairWire is also used as an in-training fairness regularizer for graph generation and link prediction. Finally, regarding our method, FAROS, we propose two variants which perform attribute switching through a uniform distribution over all the other sensitive attributes (i.e., FAROS-Uniform) and a prior distribution learned by the diffusion model (i.e., FAROS-Prior). In the Supplementary Material, we release the code to fully reproduce all results, along with complete details on the training and hyper-parameter selection for all baselines and FAROS.

\subsection{Results and discussion}
We aim to answer the following research questions (RQs): \textbf{RQ1)} How good is FAROS at striking an accuracy-fairness trade-off with respect to the other baselines for the task of link prediction? 
\textbf{RQ2)} Is the formulation for the optimal fraction of nodes to switch, $\rho^{*}$, also empirically good with respect to the settings with no switching and all-nodes-switched? \textbf{RQ3)} What is the impact on the final performance when we test other diffusion models on top of FAROS?. Besides the highlighted RQs, we also release further results for the selection of the optimal $\tau^*$ value in the Supplementary Material.

\textbf{RQ1) Overall performance.} We test the performance of FAROS against four state-of-the-art approaches and on three graph datasets for the link prediction task. Specifically, we used GraphMaker to generate 10 samples of the original graph for each dataset, trained the baselines on these samples, and tested them on the same original test set. For this reason, results are summarized (with mean and standard deviation) for all the baselines. 

\Cref{tab:results} displays the final performance results in terms of the Area Under the Curve (AUC) as the utility metric, and fairness metrics ($\Delta_{SP}$ and $\Delta_{EO}$). For each dataset, and for the sake of completeness, the first row reports the link prediction results obtained when training and testing a 1-layer Graph Autoencoder (GAE) on the original graph data. The first observable trend, and as also evidenced in~\cite{DBLP:conf/nips/KoseS24}, is that link prediction results obtained with GraphMaker are, in most cases, affected by unfairness issues. This highlights the need to mitigate the presence of bias in the generated graphs. 

In this respect, the other baselines can solve the fairness problem, but in most cases at the expense of lower accuracy values (accuracy and fairness metrics are usually known to behave in inverted directions). To this end, our approach, FAROS, is the method able to strike a better trade-off between accuracy and fairness than the other baselines, settling (in the vast majority of cases) within the top-3 performance measures for each metric. This becomes evident on \textsc{Amazon Photo} and especially \textsc{Cora}, where FAROS-Prior reaches an almost equal accuracy value to GraphMaker, while keeping low fairness. Although on \textsc{Citeseer} the observed trade-off is slightly worse compared to FairWire (our main baseline), we emphasize the strong fairness performance of FAROS, which surpasses FairWire in both the Uniform and Prior settings. 



\begin{figure}[!t]
    \centering    \includegraphics[width=0.8\columnwidth]{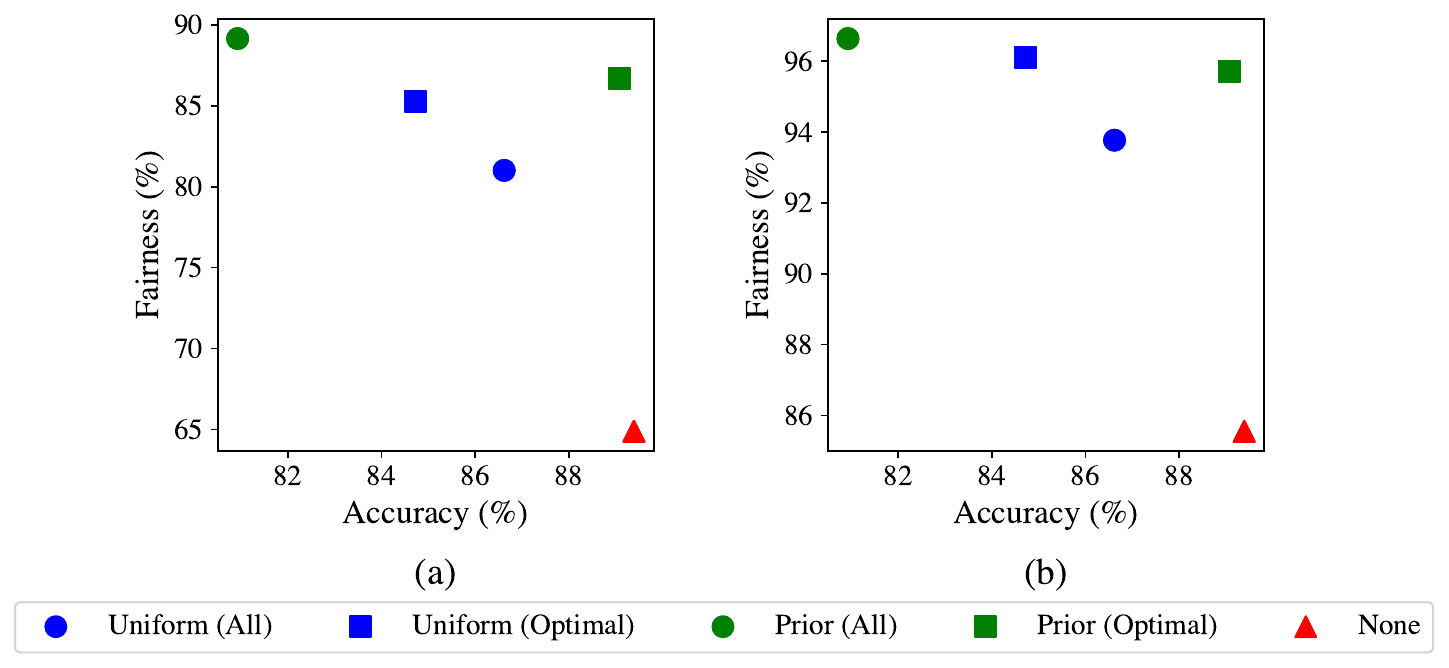} 
    \caption{Link prediction results for FAROS on \textsc{Cora} when switching no nodes (i.e., GraphMaker \protect \trianglemarker{red}), all nodes (\protect \circlemarker{blue} and \protect \circlemarker{forest}), and the optimal fraction of nodes (\protect \squaremarker{blue} and \protect \squaremarker{forest}). Accuracy is assessed using AUC, while fairness is measured in terms of $\Delta_{SP}$ (a) and $\Delta_{EO}$ (b). For better visualization, we computed 100 - value for each fairness metric to adhere to the principle ``higher is better''.}
    \label{fig:rq2}
\end{figure}

\textbf{RQ2) Calculation of the optimal switching nodes fraction $\rho^*$.} We assess the goodness of our proposed optimal fraction of nodes to be switched by comparing the results obtained in this setting with two extreme cases: no switched nodes (i.e., GraphMaker) and all switched nodes. \Cref{fig:rq2} reports the calculated performance on the \textsc{Cora} dataset, again for the two distributions, Uniform and Prior. Overall, results clearly underline how the selection of the optimal node fraction can strike a better accuracy-fairness trade-off with respect to the other two settings, empirically demonstrating the effectiveness of our formulation. 

\textbf{RQ3) Effect of different diffusion backbones.} We analyze the impact of FAROS when working on different graph diffusion model backbones. Results in \Cref{tab:diffusion_backbones} show the performance variation of FAROS on \textsc{Cora}, when backed up with GraphMaker synchronous (our default choice in this work, indicated as GraphMaker-Sync) and asynchronous versions (indicated as GraphMaker-Async), as well as the GraphMaker synchronous version trained with a fairness regularizer (i.e., FairWire). We highlight that the implementation of the attribute switching mechanisms on GraphMaker-Async slightly deviates from that proposed in the methodology section of this paper; we provide further details regarding this aspect in the Supplementary Material.

\begin{table}[!t]
\caption{Link prediction results as accuracy (AUC) and fairness ($\Delta_{SP}$ and $\Delta_{EO}$) for all the tested baselines on \textsc{Cora}, \textsc{Citeseer}, and \textsc{Amazon Photo}. For each dataset, the first row indicates the results for a GAE trained and tested on the original dataset, while the other rows represent the link prediction results for each model trained on 10 samples generated through GraphMaker and tested on the original dataset. Top-3 results are explicitly highlighted on each metric.}\label{tab:results} 
\centering 
\footnotesize
\begin{tabular}{llccc}
\toprule
\textbf{Datasets} & \textbf{Models} & AUC ($\uparrow$) & $\Delta_{SP}$ ($\downarrow$) & $\Delta_{EO}$ ($\downarrow$)
\\ \cmidrule{1-5} 
\multirow{7}{*}{\textsc{Cora}} & GAE & 94.92 & 27.71 & 11.53 \\ \cmidrule{2-5}
& GraphMaker & 89.39$\pm$0.92 \tikz[baseline=(char.base)]{
  \node[shape=circle,draw,inner sep=0.3pt] (char) {\tiny{\textbf{1}}};} & 35.12$\pm$2.19 & 14.45$\pm$0.77 \\
\cmidrule{2-5}
& FairAdj & 82.13$\pm$1.07 & 15.47$\pm$2.39 & 6.26$\pm$2.05 \\
& Adversarial & 83.66$\pm$5.64 & 16.35$\pm$9.80 & 7.82$\pm$5.84 \\
& FairWire & 83.41$\pm$7.03 & 10.03$\pm$7.59 \tikz[baseline=(char.base)]{
  \node[shape=circle,draw,inner sep=0.3pt] (char) {\tiny{\textbf{1}}};} & 4.07$\pm$2.83 \tikz[baseline=(char.base)]{
  \node[shape=circle,draw,inner sep=0.3pt] (char) {\tiny{\textbf{2}}};} \\ \cmidrule{2-5}
& FAROS-Uniform & 84.72$\pm$5.65 \tikz[baseline=(char.base)]{
  \node[shape=circle,draw,inner sep=0.3pt] (char) {\tiny{\textbf{3}}};} & 14.73$\pm$6.62 \tikz[baseline=(char.base)]{
  \node[shape=circle,draw,inner sep=0.3pt] (char) {\tiny{\textbf{3}}};} & 3.90$\pm$2.81 \tikz[baseline=(char.base)]{
  \node[shape=circle,draw,inner sep=0.3pt] (char) {\tiny{\textbf{1}}};}  \\
& FAROS-Prior & 89.08$\pm$2.72 \tikz[baseline=(char.base)]{
  \node[shape=circle,draw,inner sep=0.3pt] (char) {\tiny{\textbf{2}}};} & 13.30$\pm$8.62 \tikz[baseline=(char.base)]{
  \node[shape=circle,draw,inner sep=0.3pt] (char) {\tiny{\textbf{2}}};} & 4.30$\pm$4.03 \tikz[baseline=(char.base)]{
  \node[shape=circle,draw,inner sep=0.3pt] (char) {\tiny{\textbf{3}}};} \\
\cmidrule{1-5} 
\multirow{7}{*}{\textsc{Citeseer}} & GAE & 95.76 & 29.05 & 9.53 \\ \cmidrule{2-5}
& GraphMaker & 91.88$\pm$0.98 \tikz[baseline=(char.base)]{
  \node[shape=circle,draw,inner sep=0.3pt] (char) {\tiny{\textbf{1}}};} & 36.77$\pm$1.22 & 14.11$\pm$0.94 \\ \cmidrule{2-5}
& FairAdj & 82.67$\pm$2.78 & 15.45$\pm$2.68 \tikz[baseline=(char.base)]{
  \node[shape=circle,draw,inner sep=0.3pt] (char) {\tiny{\textbf{2}}};} & 7.98$\pm$1.47 \tikz[baseline=(char.base)]{
  \node[shape=circle,draw,inner sep=0.3pt] (char) {\tiny{\textbf{3}}};} \\
& Adversarial & 89.59$\pm$2.70 \tikz[baseline=(char.base)]{
  \node[shape=circle,draw,inner sep=0.3pt] (char) {\tiny{\textbf{3}}};} & 24.20$\pm$5.82 & 10.34$\pm$1.66 \\
& FairWire & 91.29$\pm$3.07 \tikz[baseline=(char.base)]{
  \node[shape=circle,draw,inner sep=0.3pt] (char) {\tiny{\textbf{2}}};} & 19.61$\pm$7.16 & 8.41$\pm$3.25 \\ \cmidrule{2-5}
& FAROS-Uniform & 84.45$\pm$7.71 & 12.39$\pm$8.05 \tikz[baseline=(char.base)]{
  \node[shape=circle,draw,inner sep=0.3pt] (char) {\tiny{\textbf{1}}};} & 4.94$\pm$3.30 \tikz[baseline=(char.base)]{
  \node[shape=circle,draw,inner sep=0.3pt] (char) {\tiny{\textbf{1}}};} \\
& FAROS-Prior & 87.76$\pm$6.21 & 17.94$\pm$9.46 \tikz[baseline=(char.base)]{
  \node[shape=circle,draw,inner sep=0.3pt] (char) {\tiny{\textbf{3}}};} & 7.45$\pm$3.65 \tikz[baseline=(char.base)]{
  \node[shape=circle,draw,inner sep=0.3pt] (char) {\tiny{\textbf{2}}};} \\
\cmidrule{1-5} 
\multirow{7}{*}{\textsc{Amazon Photo}} & GAE & 96.91 & 32.58 & 8.24 \\ \cmidrule{2-5}
& GraphMaker & 92.38$\pm$1.28 & 30.87$\pm$1.05 & 12.20$\pm$2.22 \\ \cmidrule{2-5}
& FairAdj & -- & -- & -- \\
& Adversarial & 94.24$\pm$1.20 \tikz[baseline=(char.base)]{
  \node[shape=circle,draw,inner sep=0.3pt] (char) {\tiny{\textbf{2}}};} & 29.17$\pm$2.83 & 7.06$\pm$2.63 \\
& FairWire & 93.16$\pm$0.32 & 24.44$\pm$1.06 \tikz[baseline=(char.base)]{
  \node[shape=circle,draw,inner sep=0.3pt] (char) {\tiny{\textbf{1}}};} & 2.56$\pm$0.47 \tikz[baseline=(char.base)]{
  \node[shape=circle,draw,inner sep=0.3pt] (char) {\tiny{\textbf{1}}};} \\ \cmidrule{2-5}
& FAROS-Uniform & 94.44$\pm$0.28 \tikz[baseline=(char.base)]{
  \node[shape=circle,draw,inner sep=0.3pt] (char) {\tiny{\textbf{1}}};} & 28.38$\pm$1.52 \tikz[baseline=(char.base)]{
  \node[shape=circle,draw,inner sep=0.3pt] (char) {\tiny{\textbf{3}}};} & 4.50$\pm$0.78 \tikz[baseline=(char.base)]{
  \node[shape=circle,draw,inner sep=0.3pt] (char) {\tiny{\textbf{3}}};} \\
& FAROS-Prior & 94.10$\pm$0.25 \tikz[baseline=(char.base)]{
  \node[shape=circle,draw,inner sep=0.3pt] (char) {\tiny{\textbf{3}}};} & 27.03$\pm$0.36 \tikz[baseline=(char.base)]{
  \node[shape=circle,draw,inner sep=0.3pt] (char) {\tiny{\textbf{2}}};} & 3.29$\pm$0.27 \tikz[baseline=(char.base)]{
  \node[shape=circle,draw,inner sep=0.3pt] (char) {\tiny{\textbf{2}}};} \\
\bottomrule
\end{tabular}
\end{table}


On the one hand, we point out that, for FairWire, our multi-criteria optimization for $\tau^*$ suggests no-switching as the configuration to reach the best trade-off; that is why the reported performance results are the same as those of FairWire in RQ1. We explain this behavior by recalling that, as observed in most cases, FairWire represents a Pareto-optimal solution, so no further intervention (such as that of FAROS) could improve its performance trade-off (refer to the Supplementary Material for all Pareto result visualizations). On the other hand, we notice that coupling GraphMaker-Async with FAROS constitutes an incredibly strong baseline, able to strike a better accuracy-fairness trade-off than the other two tested GDMs. This paves the way to explore additional GDMs on top of FAROS as future directions of the current work. 

\begin{table}[!t]
\caption{Link prediction results for FAROS on \textsc{Cora} with different diffusion backbones. }\label{tab:diffusion_backbones} 
\centering 
\footnotesize
\begin{tabular}{llccc}
\toprule
\textbf{Distributions} & \makecell[l]{\textbf{Diffusion}\\\textbf{models}} & AUC ($\uparrow$) & $\Delta_{SP}$ ($\downarrow$) & $\Delta_{EO}$ ($\downarrow$) \\
\cmidrule{1-5} 
\multirow{3}{*}{FAROS-Uniform} & GraphMaker-Sync & 84.72$\pm$5.65 & 14.73$\pm$6.62 & \textbf{3.90$\pm$2.81} \\
& GraphMaker-Async & \textbf{89.28$\pm$1.42} & 10.36$\pm$5.27 & 6.12$\pm$3.38 \\
& FairWire & 83.41$\pm$7.03 & \textbf{10.03$\pm$7.59} & 4.07$\pm$2.83 \\
\cmidrule{1-5}
\multirow{3}{*}{FAROS-Prior} & GraphMaker-Sync & 89.08$\pm$2.72 & 13.30$\pm$8.62 & 4.30$\pm$4.03 \\
& GraphMaker-Async & \textbf{90.42$\pm$1.16} & 11.67$\pm$8.76 & 6.42$\pm$4.19 \\
& FairWire & 83.41$\pm$7.03 & \textbf{10.03$\pm$7.59} & \textbf{4.07$\pm$2.83} \\
\bottomrule
\end{tabular}
\end{table}

\section{Conclusion and Limitations}
\label{sec:conclusion}
In this paper, we have proposed FAROS, a fair graph generation framework leveraging attribute switching mechanisms. Unlike previous approaches requiring a full re-training of the graph diffusion model (GDM) through fairness constraints, FAROS directly runs during the generation process of a pre-trained GDM, and works by altering nodes' sensitive attributes to ensure fairness in the generated graph. To further reduce the impact on the GDM functioning, FAROS preliminary estimates the subset of optimal nodes to perform the switching on and the optimal time step at which to run the attribute switching. Benchmarking results for link prediction demonstrate that FAROS achieves a better accuracy-fairness trade-off than other competitors, as it reduces the fairness discrepancies while keeping comparable (or even higher) accuracy performance.

\textbf{Limitations.} Despite the demonstrated effectiveness of FAROS, the model still has much room for improvement and can provide additional food for thought regarding fairness in GDM. In this sense, it could be beneficial to extend the conducted analysis to other tasks in graph machine learning (e.g., node classification) to confirm the observed trends. Similarly, we think that leveraging other GDMs from the recent literature would be important to understand if the bias amplification occurs with other GDMs and, in that case, if FAROS is still capable to handle and address the generated unfairness. In this respect, the preliminary results shown in \Cref{tab:diffusion_backbones} have demonstrated that the asynchronous version of GraphMaker can also benefit from the application of FAROS. This suggests that we should keep exploring the application of our nodes attribute switching mechanisms for a GDM where node features and topology are separately generated; indeed, our intuition is that the interplay between these two generated components in GDM could be one of the reasons behind the unfairness of these models, and directly acting on that might be beneficial. A possible limitation to the suggested direction is that the recent literature has not seen extensive research into GDMs for large attributed graphs, with the only exception of GraphMaker~\cite{DBLP:journals/corr/abs-2310-13833}. However, we acknowledge some recent works are currently exploring these ideas, so we plan to leverage such models to extend our rationale~\cite{DBLP:conf/ecai/LimniosSCMRE24, DBLP:conf/icml/TrivediRA0DKLPA24}.  

\section*{Acknowledgments}

Daniele Malitesta and Fragkiskos D. Malliaros acknowledge the support of the Innov4-ePiK project managed by the French National Research Agency under the 4th PIA, integrated into France2030 (ANR-23-RHUS-0002).

\bibliographystyle{unsrtnat}
\bibliography{references}

\begin{thebibliography}{65}
\providecommand{\natexlab}[1]{#1}
\providecommand{\url}[1]{\texttt{#1}}
\expandafter\ifx\csname urlstyle\endcsname\relax
  \providecommand{\doi}[1]{doi: #1}\else
  \providecommand{\doi}{doi: \begingroup \urlstyle{rm}\Url}\fi

\bibitem[Hamilton(2020)]{DBLP:series/synthesis/2020Hamilton}
William~L. Hamilton.
\newblock \emph{Graph Representation Learning}.
\newblock Synthesis Lectures on Artificial Intelligence and Machine Learning. Morgan {\&} Claypool Publishers, 2020.

\bibitem[He et~al.(2020)He, Deng, Wang, Li, Zhang, and Wang]{DBLP:conf/sigir/0001DWLZ020}
Xiangnan He, Kuan Deng, Xiang Wang, Yan Li, Yong{-}Dong Zhang, and Meng Wang.
\newblock {LightGCN}: Simplifying and powering graph convolution network for recommendation.
\newblock In \emph{{SIGIR}}, 2020.

\bibitem[Cai et~al.(2023)Cai, Huang, Xia, and Ren]{DBLP:conf/iclr/Cai0XR23}
Xuheng Cai, Chao Huang, Lianghao Xia, and Xubin Ren.
\newblock {LightGCL}: Simple yet effective graph contrastive learning for recommendation.
\newblock In \emph{{ICLR}}, 2023.

\bibitem[Ezzat et~al.(2017)Ezzat, Zhao, Wu, Li, and Kwoh]{DBLP:journals/tcbb/EzzatZW0K17}
Ali Ezzat, Peilin Zhao, Min Wu, Xiaoli Li, and Chee~Keong Kwoh.
\newblock Drug-target interaction prediction with graph regularized matrix factorization.
\newblock \emph{{IEEE} {ACM} Trans. Comput. Biol. Bioinform.}, 14\penalty0 (3):\penalty0 646--656, 2017.

\bibitem[Ning et~al.(2025)Ning, Wang, Zhao, Sun, Jiang, Wang, and Yin]{DBLP:journals/artmed/NingWZSJWY25}
Qiao Ning, Yue Wang, Yaomiao Zhao, Jiahao Sun, Lu~Jiang, Kaidi Wang, and Minghao Yin.
\newblock {DMHGNN:} double multi-view heterogeneous graph neural network framework for drug-target interaction prediction.
\newblock \emph{Artif. Intell. Medicine}, 159:\penalty0 103023, 2025.

\bibitem[Castro-Correa et~al.(2024)Castro-Correa, Giraldo, Badiey, and Malliaros]{gegenGNN-TNNLS24}
Jhon~A. Castro-Correa, Jhony~H. Giraldo, Mohsen Badiey, and Fragkiskos~D. Malliaros.
\newblock Gegenbauer graph neural networks for time-varying signal reconstruction.
\newblock \emph{IEEE Transactions on Neural Networks and Learning Systems}, 35\penalty0 (9):\penalty0 11734--11745, 2024.

\bibitem[Scarselli et~al.(2009)Scarselli, Gori, Tsoi, Hagenbuchner, and Monfardini]{DBLP:journals/tnn/ScarselliGTHM09}
Franco Scarselli, Marco Gori, Ah~Chung Tsoi, Markus Hagenbuchner, and Gabriele Monfardini.
\newblock The graph neural network model.
\newblock \emph{{IEEE} Trans. Neural Networks}, 20\penalty0 (1):\penalty0 61--80, 2009.

\bibitem[Yang et~al.(2023{\natexlab{a}})Yang, Wu, Wang, and Yan]{DBLP:conf/iclr/YangWWY23}
Chenxiao Yang, Qitian Wu, Jiahua Wang, and Junchi Yan.
\newblock Graph neural networks are inherently good generalizers: Insights by bridging {GNNs} and {MLPs}.
\newblock In \emph{{ICLR}}, 2023{\natexlab{a}}.

\bibitem[Caton and Haas(2024)]{DBLP:journals/csur/CatonH24}
Simon Caton and Christian Haas.
\newblock Fairness in machine learning: {A} survey.
\newblock \emph{{ACM} Comput. Surv.}, 56\penalty0 (7):\penalty0 166:1--166:38, 2024.

\bibitem[Dai and Wang(2021)]{DBLP:conf/wsdm/DaiW21}
Enyan Dai and Suhang Wang.
\newblock Say no to the discrimination: Learning fair graph neural networks with limited sensitive attribute information.
\newblock In \emph{{WSDM}}, 2021.

\bibitem[Li et~al.(2021)Li, Wang, Zhao, Hong, and Liu]{DBLP:conf/iclr/LiWZHL21}
Peizhao Li, Yifei Wang, Han Zhao, Pengyu Hong, and Hongfu Liu.
\newblock On dyadic fairness: Exploring and mitigating bias in graph connections.
\newblock In \emph{{ICLR}}, 2021.

\bibitem[Spinelli et~al.(2022)Spinelli, Scardapane, Hussain, and Uncini]{DBLP:journals/tai/SpinelliSHU22}
Indro Spinelli, Simone Scardapane, Amir Hussain, and Aurelio Uncini.
\newblock Fairdrop: Biased edge dropout for enhancing fairness in graph representation learning.
\newblock \emph{{IEEE} Trans. Artif. Intell.}, 3\penalty0 (3):\penalty0 344--354, 2022.

\bibitem[K{\"{o}}se and Shen(2024)]{DBLP:journals/tkdd/KoseS24}
{\"{O}}yk{\"{u}}~Deniz K{\"{o}}se and Yanning Shen.
\newblock Fairgat: Fairness-aware graph attention networks.
\newblock \emph{{ACM} Trans. Knowl. Discov. Data}, 18\penalty0 (7):\penalty0 164, 2024.

\bibitem[Wang et~al.(2025)Wang, Chu, Doan, Wang, Wu, Palade, and Zhang]{DBLP:conf/aaai/WangCDWWPZ25}
Zichong Wang, Zhibo Chu, Thang~Viet Doan, Shaowei Wang, Yongkai Wu, Vasile Palade, and Wenbin Zhang.
\newblock Fair graph {U-Net}: {A} fair graph learning framework integrating group and individual awareness.
\newblock In \emph{{AAAI}}, 2025.

\bibitem[Zehlike et~al.(2017)Zehlike, Bonchi, Castillo, Hajian, Megahed, and Baeza{-}Yates]{DBLP:conf/cikm/ZehlikeB0HMB17}
Meike Zehlike, Francesco Bonchi, Carlos Castillo, Sara Hajian, Mohamed Megahed, and Ricardo Baeza{-}Yates.
\newblock {FA*IR:} {A} fair top-k ranking algorithm.
\newblock In \emph{{CIKM}}, 2017.

\bibitem[Chen et~al.(2024)Chen, Rossi, Park, Trivedi, Wang, Yu, Kim, Dernoncourt, and Ahmed]{DBLP:journals/tkdd/ChenRPTWYKDA24}
April Chen, Ryan~A. Rossi, Namyong Park, Puja Trivedi, Yu~Wang, Tong Yu, Sungchul Kim, Franck Dernoncourt, and Nesreen~K. Ahmed.
\newblock Fairness-aware graph neural networks: {A} survey.
\newblock \emph{{ACM} Trans. Knowl. Discov. Data}, 18\penalty0 (6):\penalty0 138:1--138:23, 2024.

\bibitem[Choudhary et~al.(2022)Choudhary, Laclau, and Largeron]{DBLP:journals/corr/abs-2205-05396}
Manvi Choudhary, Charlotte Laclau, and Christine Largeron.
\newblock A survey on fairness for machine learning on graphs.
\newblock \emph{CoRR}, abs/2205.05396, 2022.

\bibitem[Dong et~al.(2023)Dong, Ma, Wang, Chen, and Li]{DBLP:journals/tkde/DongMWCL23}
Yushun Dong, Jing Ma, Song Wang, Chen Chen, and Jundong Li.
\newblock Fairness in graph mining: {A} survey.
\newblock \emph{{IEEE} Trans. Knowl. Data Eng.}, 35\penalty0 (10):\penalty0 10583--10602, 2023.

\bibitem[Cao et~al.(2024)Cao, Tan, Gao, Xu, Chen, Heng, and Li]{DBLP:journals/tkde/CaoTGXCHL24}
Hanqun Cao, Cheng Tan, Zhangyang Gao, Yilun Xu, Guangyong Chen, Pheng{-}Ann Heng, and Stan~Z. Li.
\newblock A survey on generative diffusion models.
\newblock \emph{{IEEE} Trans. Knowl. Data Eng.}, 36\penalty0 (7):\penalty0 2814--2830, 2024.

\bibitem[Xing et~al.(2025)Xing, Feng, Chen, Dai, Hu, Xu, Wu, and Jiang]{DBLP:journals/csur/XingFCDHXWJ25}
Zhen Xing, Qijun Feng, Haoran Chen, Qi~Dai, Han Hu, Hang Xu, Zuxuan Wu, and Yu{-}Gang Jiang.
\newblock A survey on video diffusion models.
\newblock \emph{{ACM} Comput. Surv.}, 57\penalty0 (2):\penalty0 41:1--41:42, 2025.

\bibitem[Kingma and Welling(2014)]{DBLP:journals/corr/KingmaW13}
Diederik~P. Kingma and Max Welling.
\newblock Auto-encoding variational bayes.
\newblock In \emph{{ICLR}}, 2014.

\bibitem[Goodfellow et~al.(2014)Goodfellow, Pouget{-}Abadie, Mirza, Xu, Warde{-}Farley, Ozair, Courville, and Bengio]{DBLP:journals/corr/GoodfellowPMXWOCB14}
Ian~J. Goodfellow, Jean Pouget{-}Abadie, Mehdi Mirza, Bing Xu, David Warde{-}Farley, Sherjil Ozair, Aaron~C. Courville, and Yoshua Bengio.
\newblock Generative adversarial networks.
\newblock \emph{CoRR}, abs/1406.2661, 2014.

\bibitem[Chen et~al.(2025)Chen, Xiang, Hu, Ye, Yu, Cheng, and Zhang]{DBLP:journals/air/ChenXHYYCZ25}
Hang Chen, Qian Xiang, Jiaxin Hu, Meilin Ye, Chao Yu, Hao Cheng, and Lei Zhang.
\newblock Comprehensive exploration of diffusion models in image generation: a survey.
\newblock \emph{Artif. Intell. Rev.}, 58\penalty0 (4):\penalty0 99, 2025.

\bibitem[Ramesh et~al.(2021)Ramesh, Pavlov, Goh, Gray, Voss, Radford, Chen, and Sutskever]{DBLP:conf/icml/RameshPGGVRCS21}
Aditya Ramesh, Mikhail Pavlov, Gabriel Goh, Scott Gray, Chelsea Voss, Alec Radford, Mark Chen, and Ilya Sutskever.
\newblock Zero-shot text-to-image generation.
\newblock In \emph{{ICML}}, 2021.

\bibitem[Rombach et~al.(2022)Rombach, Blattmann, Lorenz, Esser, and Ommer]{DBLP:conf/cvpr/RombachBLEO22}
Robin Rombach, Andreas Blattmann, Dominik Lorenz, Patrick Esser, and Bj{\"{o}}rn Ommer.
\newblock High-resolution image synthesis with latent diffusion models.
\newblock In \emph{{CVPR}}, 2022.

\bibitem[Child(2021)]{DBLP:conf/iclr/Child21}
Rewon Child.
\newblock Very deep {VAEs} generalize autoregressive models and can outperform them on images.
\newblock In \emph{{ICLR}}, 2021.

\bibitem[Saharia et~al.(2023)Saharia, Ho, Chan, Salimans, Fleet, and Norouzi]{DBLP:journals/pami/SahariaHCSFN23}
Chitwan Saharia, Jonathan Ho, William Chan, Tim Salimans, David~J. Fleet, and Mohammad Norouzi.
\newblock Image super-resolution via iterative refinement.
\newblock \emph{{IEEE} Trans. Pattern Anal. Mach. Intell.}, 45\penalty0 (4):\penalty0 4713--4726, 2023.

\bibitem[Han et~al.(2023)Han, Kumar, and Tsvetkov]{DBLP:conf/acl/HanKT23}
Xiaochuang Han, Sachin Kumar, and Yulia Tsvetkov.
\newblock {SSD-LM:} semi-autoregressive simplex-based diffusion language model for text generation and modular control.
\newblock In \emph{{ACL}}, 2023.

\bibitem[Li et~al.(2022)Li, Thickstun, Gulrajani, Liang, and Hashimoto]{DBLP:conf/nips/LiTGLH22}
Xiang~Lisa Li, John Thickstun, Ishaan Gulrajani, Percy Liang, and Tatsunori~B. Hashimoto.
\newblock Diffusion-lm improves controllable text generation.
\newblock In \emph{NeurIPS}, 2022.

\bibitem[Ramesh et~al.(2022)Ramesh, Dhariwal, Nichol, Chu, and Chen]{DBLP:journals/corr/abs-2204-06125}
Aditya Ramesh, Prafulla Dhariwal, Alex Nichol, Casey Chu, and Mark Chen.
\newblock Hierarchical text-conditional image generation with {CLIP} latents.
\newblock \emph{CoRR}, abs/2204.06125, 2022.

\bibitem[Yang et~al.(2024)Yang, Yu, Meng, Xu, Ermon, and Cui]{DBLP:conf/icml/0006YMXE024}
Ling Yang, Zhaochen Yu, Chenlin Meng, Minkai Xu, Stefano Ermon, and Bin Cui.
\newblock Mastering text-to-image diffusion: Recaptioning, planning, and generating with multimodal {LLMs}.
\newblock In \emph{{ICML}}, 2024.

\bibitem[Liu et~al.(2023)Liu, Fan, Liu, Li, Li, Liu, Tang, and Li]{DBLP:conf/ijcai/LiuFLLLLTL23}
Chengyi Liu, Wenqi Fan, Yunqing Liu, Jiatong Li, Hang Li, Hui Liu, Jiliang Tang, and Qing Li.
\newblock Generative diffusion models on graphs: Methods and applications.
\newblock In \emph{{IJCAI}}, 2023.

\bibitem[Zhu et~al.(2022)Zhu, Du, Wang, Xu, Zhang, Liu, and Wu]{DBLP:conf/log/0001DWXZLW22}
Yanqiao Zhu, Yuanqi Du, Yinkai Wang, Yichen Xu, Jieyu Zhang, Qiang Liu, and Shu Wu.
\newblock A survey on deep graph generation: Methods and applications.
\newblock In \emph{LoG}, 2022.

\bibitem[Guo and Zhao(2023)]{DBLP:journals/pami/GuoZ23}
Xiaojie Guo and Liang Zhao.
\newblock A systematic survey on deep generative models for graph generation.
\newblock \emph{{IEEE} Trans. Pattern Anal. Mach. Intell.}, 45\penalty0 (5):\penalty0 5370--5390, 2023.

\bibitem[Hoogeboom et~al.(2022)Hoogeboom, Satorras, Vignac, and Welling]{DBLP:conf/icml/HoogeboomSVW22}
Emiel Hoogeboom, Victor~Garcia Satorras, Cl{\'{e}}ment Vignac, and Max Welling.
\newblock Equivariant diffusion for molecule generation in 3d.
\newblock In \emph{{ICML}}, 2022.

\bibitem[Xu et~al.(2022)Xu, Yu, Song, Shi, Ermon, and Tang]{DBLP:conf/iclr/XuY0SE022}
Minkai Xu, Lantao Yu, Yang Song, Chence Shi, Stefano Ermon, and Jian Tang.
\newblock {GeoDiff:} {A} geometric diffusion model for molecular conformation generation.
\newblock In \emph{{ICLR}}, 2022.

\bibitem[Zhang et~al.(2023)Zhang, Qamar, Kang, Jung, Zhang, Bae, and Zhang]{DBLP:journals/corr/abs-2304-01565}
Mengchun Zhang, Maryam Qamar, Taegoo Kang, Yuna Jung, Chenshuang Zhang, Sung{-}Ho Bae, and Chaoning Zhang.
\newblock A survey on graph diffusion models: Generative {AI} in science for molecule, protein and material.
\newblock \emph{CoRR}, abs/2304.01565, 2023.

\bibitem[Yang et~al.(2023{\natexlab{b}})Yang, Yang, Zhou, and Sun]{DBLP:conf/nips/YangYZS23}
Run Yang, Yuling Yang, Fan Zhou, and Qiang Sun.
\newblock Directional diffusion models for graph representation learning.
\newblock In \emph{NeurIPS}, 2023{\natexlab{b}}.

\bibitem[Jo et~al.(2022)Jo, Lee, and Hwang]{DBLP:conf/icml/JoLH22}
Jaehyeong Jo, Seul Lee, and Sung~Ju Hwang.
\newblock Score-based generative modeling of graphs via the system of stochastic differential equations.
\newblock In \emph{{ICML}}, 2022.

\bibitem[Niu et~al.(2020)Niu, Song, Song, Zhao, Grover, and Ermon]{DBLP:conf/aistats/NiuSSZGE20}
Chenhao Niu, Yang Song, Jiaming Song, Shengjia Zhao, Aditya Grover, and Stefano Ermon.
\newblock Permutation invariant graph generation via score-based generative modeling.
\newblock In \emph{{AISTATS}}, 2020.

\bibitem[Huang et~al.(2022)Huang, Sun, Du, Fu, and Lv]{DBLP:conf/icdm/HuangS0FL22}
Han Huang, Leilei Sun, Bowen Du, Yanjie Fu, and Weifeng Lv.
\newblock {GraphGDP}: Generative diffusion processes for permutation invariant graph generation.
\newblock In \emph{{ICDM}}, pages 201--210. {IEEE}, 2022.

\bibitem[Vignac et~al.(2023)Vignac, Krawczuk, Siraudin, Wang, Cevher, and Frossard]{DBLP:conf/iclr/VignacKSWCF23}
Cl{\'{e}}ment Vignac, Igor Krawczuk, Antoine Siraudin, Bohan Wang, Volkan Cevher, and Pascal Frossard.
\newblock Digress: Discrete denoising diffusion for graph generation.
\newblock In \emph{{ICLR}}, 2023.

\bibitem[Xu et~al.(2024)Xu, Qiu, Chen, Chen, Fan, Pan, Zeng, Das, and Tong]{DBLP:conf/nips/0007QCCFP0DT24}
Zhe Xu, Ruizhong Qiu, Yuzhong Chen, Huiyuan Chen, Xiran Fan, Menghai Pan, Zhichen Zeng, Mahashweta Das, and Hanghang Tong.
\newblock Discrete-state continuous-time diffusion for graph generation.
\newblock In \emph{NeurIPS}, 2024.

\bibitem[Liu et~al.(2024)Liu, Xu, Luo, and Jiang]{DBLP:journals/corr/abs-2401-13858}
Gang Liu, Jiaxin Xu, Tengfei Luo, and Meng Jiang.
\newblock Inverse molecular design with multi-conditional diffusion guidance.
\newblock \emph{CoRR}, abs/2401.13858, 2024.

\bibitem[Siraudin et~al.(2025)Siraudin, Malliaros, and Morris]{siraudin2025cometh}
Antoine Siraudin, Fragkiskos~D. Malliaros, and Christopher Morris.
\newblock Cometh: A continuous-time discrete-state graph diffusion model.
\newblock \emph{Transactions on Machine Learning Research}, 2025.

\bibitem[Li et~al.(2024)Li, Kreacic, Potluru, and Li]{DBLP:journals/corr/abs-2310-13833}
Mufei Li, Eleonora Kreacic, Vamsi~K. Potluru, and Pan Li.
\newblock {GraphMaker}: Can diffusion models generate large attributed graphs?
\newblock \emph{Transactions on Machine Learning Research}, 2024.

\bibitem[Trivedi et~al.(2024)Trivedi, Rossi, Arbour, Yu, Dernoncourt, Kim, Lipka, Park, Ahmed, and Koutra]{DBLP:conf/icml/TrivediRA0DKLPA24}
Puja Trivedi, Ryan~A. Rossi, David Arbour, Tong Yu, Franck Dernoncourt, Sungchul Kim, Nedim Lipka, Namyong Park, Nesreen~K. Ahmed, and Danai Koutra.
\newblock Editing partially observable networks via graph diffusion models.
\newblock In \emph{{ICML}}, 2024.

\bibitem[Friedrich et~al.(2023)Friedrich, Schramowski, Brack, Struppek, Hintersdorf, Luccioni, and Kersting]{DBLP:journals/corr/abs-2302-10893}
Felix Friedrich, Patrick Schramowski, Manuel Brack, Lukas Struppek, Dominik Hintersdorf, Sasha Luccioni, and Kristian Kersting.
\newblock Fair diffusion: Instructing text-to-image generation models on fairness.
\newblock \emph{CoRR}, abs/2302.10893, 2023.

\bibitem[Bianchi et~al.(2023)Bianchi, Kalluri, Durmus, Ladhak, Cheng, Nozza, Hashimoto, Jurafsky, Zou, and Caliskan]{DBLP:conf/fat/0001KDLCNHJ0C23}
Federico Bianchi, Pratyusha Kalluri, Esin Durmus, Faisal Ladhak, Myra Cheng, Debora Nozza, Tatsunori Hashimoto, Dan Jurafsky, James Zou, and Aylin Caliskan.
\newblock Easily accessible text-to-image generation amplifies demographic stereotypes at large scale.
\newblock In \emph{FAccT}, 2023.

\bibitem[Cho et~al.(2023)Cho, Zala, and Bansal]{DBLP:conf/iccv/0001ZB23}
Jaemin Cho, Abhay Zala, and Mohit Bansal.
\newblock {DALL-EVAL:} probing the reasoning skills and social biases of text-to-image generation models.
\newblock In \emph{{ICCV}}, 2023.

\bibitem[Malitesta et~al.(2024)Malitesta, Medda, Purificato, Boratto, Malliaros, Marras, and Luca]{DBLP:journals/corr/abs-2409-04339}
Daniele Malitesta, Giacomo Medda, Erasmo Purificato, Ludovico Boratto, Fragkiskos~D. Malliaros, Mirko Marras, and Ernesto William~De Luca.
\newblock How fair is your diffusion recommender model?
\newblock \emph{CoRR}, abs/2409.04339, 2024.

\bibitem[Choi et~al.(2024)Choi, Park, Kim, Lee, and Park]{DBLP:conf/aaai/Choi0K0P24}
Yujin Choi, Jinseong Park, Hoki Kim, Jaewook Lee, and Saerom Park.
\newblock Fair sampling in diffusion models through switching mechanism.
\newblock In \emph{{AAAI}}, 2024.

\bibitem[Shen et~al.(2023)Shen, Du, Pang, Lin, Wong, and Kankanhalli]{DBLP:journals/corr/abs-2311-07604}
Xudong Shen, Chao Du, Tianyu Pang, Min Lin, Yongkang Wong, and Mohan~S. Kankanhalli.
\newblock Finetuning text-to-image diffusion models for fairness.
\newblock \emph{CoRR}, abs/2311.07604, 2023.

\bibitem[Kang et~al.(2025)Kang, Kumar, Roy, Kumar, Khosla, Narayanaswamy, and Gangadharaiah]{DBLP:journals/corr/abs-2503-01872}
Mintong Kang, Vinayshekhar~Bannihatti Kumar, Shamik Roy, Abhishek Kumar, Sopan Khosla, Balakrishnan Narayanaswamy, and Rashmi Gangadharaiah.
\newblock {FairGen}: Controlling sensitive attributes for fair generations in diffusion models via adaptive latent guidance.
\newblock \emph{CoRR}, abs/2503.01872, 2025.

\bibitem[Navarro et~al.(2024{\natexlab{a}})Navarro, Rey, Buciulea, Marques, and Segarra]{DBLP:conf/eusipco/NavarroRBMS24}
Madeline Navarro, Samuel Rey, Andrei Buciulea, Antonio~G. Marques, and Santiago Segarra.
\newblock Mitigating subpopulation bias for fair network topology inference.
\newblock In \emph{{EUSIPCO}}, 2024{\natexlab{a}}.

\bibitem[Navarro et~al.(2024{\natexlab{b}})Navarro, Rey, Buciulea, Marques, and Segarra]{DBLP:conf/nips/NavarroRBMS24}
Madeline Navarro, Samuel Rey, Andrei Buciulea, Antonio~G. Marques, and Santiago Segarra.
\newblock Fair {GLASSO:} estimating fair graphical models with unbiased statistical behavior.
\newblock In \emph{NeurIPS}, 2024{\natexlab{b}}.

\bibitem[Zheng et~al.(2024)Zheng, Zhou, Tong, Xu, Zhu, and He]{DBLP:conf/icde/Zheng0TXZH24}
Lecheng Zheng, Dawei Zhou, Hanghang Tong, Jiejun Xu, Yada Zhu, and Jingrui He.
\newblock {FairGen}: Towards fair graph generation.
\newblock In \emph{{ICDE}}, 2024.

\bibitem[Kose and Shen(2024)]{DBLP:conf/nips/KoseS24}
Oyku~Deniz Kose and Yanning Shen.
\newblock {FairWire}: Fair graph generation.
\newblock In \emph{NeurIPS}, 2024.

\bibitem[Vayer et~al.(2019)Vayer, Courty, Tavenard, Chapel, and Flamary]{DBLP:conf/icml/VayerCTCF19}
Titouan Vayer, Nicolas Courty, Romain Tavenard, Laetitia Chapel, and R{\'{e}}mi Flamary.
\newblock Optimal transport for structured data with application on graphs.
\newblock In \emph{{ICML}}, 2019.

\bibitem[Sen et~al.(2008)Sen, Namata, Bilgic, Getoor, Gallagher, and Eliassi{-}Rad]{DBLP:journals/aim/SenNBGGE08}
Prithviraj Sen, Galileo Namata, Mustafa Bilgic, Lise Getoor, Brian Gallagher, and Tina Eliassi{-}Rad.
\newblock Collective classification in network data.
\newblock \emph{{AI} Mag.}, 29\penalty0 (3):\penalty0 93--106, 2008.

\bibitem[Shchur et~al.(2018)Shchur, Mumme, Bojchevski, and G{\"{u}}nnemann]{DBLP:journals/corr/abs-1811-05868}
Oleksandr Shchur, Maximilian Mumme, Aleksandar Bojchevski, and Stephan G{\"{u}}nnemann.
\newblock Pitfalls of graph neural network evaluation.
\newblock \emph{CoRR}, abs/1811.05868, 2018.

\bibitem[Limnios et~al.(2024)Limnios, Selvaraj, Cucuringu, Maple, Reinert, and Elliott]{DBLP:conf/ecai/LimniosSCMRE24}
Stratis Limnios, Praveen Selvaraj, Mihai Cucuringu, Carsten Maple, Gesine Reinert, and Andrew Elliott.
\newblock Sagess: {A} sampling graph denoising diffusion model for scalable graph generation.
\newblock In \emph{{ECAI}}, 2024.

\bibitem[Villani(2008)]{Wasserstein}
Cédric Villani.
\newblock \emph{Optimal Transport: Old and New}.
\newblock Grundlehren der mathematischen Wissenschaften. Springer Berlin, Heidelberg, 2008.

\bibitem[M{\'{e}}moli(2011)]{DBLP:journals/focm/Memoli11}
Facundo M{\'{e}}moli.
\newblock Gromov-wasserstein distances and the metric approach to object matching.
\newblock \emph{Found. Comput. Math.}, 11\penalty0 (4):\penalty0 417--487, 2011.

\bibitem[Solomon et~al.(2016)Solomon, Peyr{\'{e}}, Kim, and Sra]{DBLP:journals/tog/SolomonPKS16}
Justin Solomon, Gabriel Peyr{\'{e}}, Vladimir~G. Kim, and Suvrit Sra.
\newblock Entropic metric alignment for correspondence problems.
\newblock \emph{{ACM} Trans. Graph.}, 35\penalty0 (4):\penalty0 72:1--72:13, 2016.

\end{thebibliography}

\newpage

\appendix

\section*{\Large Supplementary Material}

The Supplementary Material is organized as follows. First, in \Cref{sec:add_background}, we provide the demonstration for the optimal selection of $\rho^*$ and further theoretical details about the Fused Gromov-Wasserstein distance in FAROS. Secondly, in \Cref{sec:add_settings}, we report on the complete pipeline we followed to conduct our experiments in terms of datasets, baselines, hyper-parameters, training/evaluation pipelines, and computational resources. Then, in \Cref{sec:add_experiments}, we present additional results for the Pareto optimality visualization, the calculated optimal $\rho^*$ values and $\tau^*$ selection, complementary experiments with other diffusion models on \textsc{Amazon Photo}, and topological differences between original and generated graphs. Finally, in \Cref{sec:broader_impact}, we describe the possible broader impacts of our work.

\section{Additional Details on the Theoretical Formulation of FAROS}
\label{sec:add_background}

\subsection{Demonstration of optimal $\rho^*$ selection}

In this section, we provide a formal demonstration on how we can estimate the optimal fraction of nodes to be switched $\rho^* = \mathbb{E}[|\mathcal{V}^*| / |\mathcal{V}|]$, where $\mathcal{V}^* \subseteq \mathcal{V}$ is the set of such nodes (refer to \Cref{sec:rho_estimation}). As mentioned in the paper, our intuition is that the GDM generation amplifies the imbalance between exterior links (those connecting nodes with different sensitive attributes) and interior links (those connecting nodes with the same sensitive attributes). Thus, our purpose is to estimate the fraction of nodes to be switched such that the generated imbalance is counteracted.


\textbf{Assumption 1.} Our first assumption for this demonstration is that the pre-trained GDM perfectly mimics the original graph prior distribution of edges across sensitive attributes. Motivated by this assumption, and for the sake of simplicity, we will use the terminology ``original graph'' and ``generated graph without attribute switching'' interchangeably.

\textbf{Assumption 2.} Secondly, we assume that the existence of a link between two nodes depends only on their respective sensitive attributes.

\textbf{Notations.} Let $\mathcal{E}_{\text{int}}$, $\mathcal{E}_{\text{ext}}$ and $\tilde{\mathcal{E}}_{\text{int}}$, $\tilde{\mathcal{E}}_{\text{ext}}$ be the set of edges for interior and exterior links of the original and generated (with attribute switching) graphs, respectively. Moreover, let the following random variables be defined: $\tilde{Y}_{ij}$, an indicator variable for the existence of a link between node $i$ and node $j$ in the graph generated using the attribute switching mechanism\footnote{In general, for the referenced notation, nodes $i$ and $j$ may or may not be affected by switching.}; $Y_{ij}$, an indicator variable for the existence of a link between node $i$ and node $j$ in the generated graph without attribute switching; $N^{k \rightarrow l}$, the number of nodes of a graph generated using the attribute switching mechanism, whose sensitive attribute was switched from sensitive attribute $k$ to sensitive attribute $l$; $\tilde{E}^{(l,k)}$, a random variable representing, in the generated graph with switching, the number of actual edges between sensitive attributes $l$ and $k$ over the number of possible edges between sensitive attributes $l$ and $k$; $E^{(l,k)}$, a random variable representing, in the generated graph without switching, the number of actual edges between sensitive attributes $l$ and $k$ over the number of possible edges between sensitive attributes $l$ and $k$; $S_{i}$, the real sensitive attribute of node $i$; $Z_{i}$, the sensitive attribute of node $i$ after switching. We recall that $N$ is the total number of nodes and $\mathcal{S} = \{s_1, s_2, \ldots, s_K\}$, with $|\mathcal{S}| = K$ the total number of sensitive attribute classes. Moreover, we consider that, in the beginning of the generation, the sampled sensitive attributes are the same for both generation with and without switching.  

\textbf{Interior links.} The total number of interior edges of the generated graph after switching, $|\tilde{\mathcal{E}}_{\text{int}}|$, is the sum of indicators $\tilde{Y}_{ij}$ where both connected nodes $i$ and $j$ have the same real sensitive attribute $S_i = S_j$. We can decompose this sum based on whether the link already existed before switching or has been specifically affected by switching:
\begin{equation}
|\tilde{\mathcal{E}}_{\text{int}}| = \sum_{i=1}^{N}\sum_{j=1}^{N}\tilde{Y}_{ij}\mathbb{I}_{\{S_{i}=S_{j}\}} = \sum_{i=1}^{N}\sum_{j=1}^{N}Y_{ij}\mathbb{I}_{\{S_{i}=S_{j}\}} + \sum_{i=1}^{N}\sum_{j=1}^{N}(\tilde{Y}_{ij}-Y_{ij})\mathbb{I}_{\{S_{i}=S_{j}\}},
\end{equation}
where $\mathbb{I}$ is the indicator function equal to 1 if the condition is true, otherwise 0. Let us focus on the last contribution from the above equation, to be interpreted as the difference term representing the change in the number of interior edges due to switching ($\Delta_{\text{int}}$):
\begin{equation}
\Delta_{\text{int}} = \sum_{i=1}^{N}\sum_{j=1}^{N}(\tilde{Y}_{ij}-Y_{ij})\mathbb{I}_{\{S_{i}=S_{j}\}}.
\end{equation}
We can further decompose the indicator $\mathbb{I}_{\{S_{i}=S_{j}\}}$ based on whether we performed switching on nodes $i$ and $j$ ($Z_i \ne S_i$ or $Z_j \ne S_j$) or not ($Z_i = S_i$ or $Z_j = S_j$):
\begin{align}
\begin{split}
 \Delta_{\text{int}} &= \sum_{i=1}^{N}\sum_{j=1}^{N}(\tilde{Y}_{ij}-Y_{ij})\mathbb{I}_{\{S_{i}=S_{j}\}} \left(\mathbb{I}_{\{Z_{i}=S_{i},Z_{j}=S_{j}\}} + \mathbb{I}_{\{Z_{i}\ne S_{i},Z_{j}=S_{j}\}} + \mathbb{I}_{\{Z_{i}=S_{i},Z_{j}\ne S_{j}\}} + \mathbb{I}_{\{Z_{i}\ne S_{i},Z_{j}\ne S_{j}\}}\right) \\
    &= \sum_{i=1}^{N}\sum_{j=1}^{N}(\tilde{Y}_{ij}-Y_{ij})\left(\mathbb{I}_{\{S_{i}=S_{j}=Z_{i}=Z_{j}\}} + \mathbb{I}_{\{Z_{i}=S_{i}=S_{j},Z_{j}\ne S_{j}\}} + \mathbb{I}_{\{S_{i}=S_{j}=Z_{j},Z_{i}\ne S_{i}\}} + \mathbb{I}_{\{S_{i}=S_{j},Z_{i}\ne S_{i},Z_{j}\ne S_{j}\}}\right) \\
    & \text{(by symmetry)}\\
    &= \sum_{i=1}^{N}\sum_{j=1}^{N}(\tilde{Y}_{ij}-Y_{ij})\left(\mathbb{I}_{\{S_{i}=S_{j}=Z_{i}=Z_{j}\}} + 2\mathbb{I}_{\{Z_{i}=S_{i}=S_{j},Z_{j}\ne S_{j}\}} + \mathbb{I}_{\{S_{i}=S_{j},Z_{i}\ne S_{i},Z_{j}\ne S_{j}\}}\right).
\end{split}
\end{align}
Let us separately consider the three conditions outlined in the previous equation, namely, $\mathbb{I}_{\{S_{i}=S_{j}=Z_{i}=Z_{j}\}}$, $\mathbb{I}_{\{Z_{i}=S_{i}=S_{j},Z_{j}\ne S_{j}\}}$, and $\mathbb{I}_{\{S_{i}=S_{j},Z_{i}\ne S_{i},Z_{j}\ne S_{j}\}}$.

First, if both end nodes did not switch their original attributes ($Z_i=S_i, Z_j=S_j$), the link probability is the same as the prior (according to our \textbf{Assumption 1}), thus leading to no expected change:
\begin{equation}
\label{eq:first_term}
\mathbb{E}\left[\sum_{i=1}^{N}\sum_{j=1}^{N}(\tilde{Y}_{ij}-Y_{ij})\mathbb{I}_{\{S_{i}=S_{j}=Z_{i}=Z_{j}\}}\right] = 0 \quad \text{\textbf{(Assumption 1)}}.
\end{equation}

Second, let us consider the case where only one node was switched. Let $S_i=S_j=l$ be the original attribute, and $Z_j=k$ where $k \ne l$ is the new attribute of node $j$, while $Z_i=S_i=l$. This corresponds to a node originally in class $l$ switching to class $k$. The sum can be rewritten by summing over the original class $l$ and the new class $k$:
\begin{align}
\begin{split}
& \sum_{i=1}^{N}\sum_{j=1}^{N}(\tilde{Y}_{ij}-Y_{ij})\mathbb{I}_{\{Z_{i}=S_{i}=S_{j},Z_{j}\ne S_{j}\}} \\
    &= \sum_{l=1}^{K}\sum_{k=1; k\neq l}^{K}\sum_{i,j}(\tilde{Y}_{ij}-Y_{ij})\mathbb{I}_{\{S_{i}=l, S_{j}=l, Z_{i}=l, Z_{j}=k\}} \\
    &=\sum_{l=1}^{K}\sum_{k=1, k\ne l}^{K} N^{l \rightarrow k} N^{l \rightarrow l} (\tilde{E}^{(l,k)}-E^{(l,l)}) \quad \text{\textbf{(Assumption 2)}}.
\end{split}
\end{align}
This equality holds as a link between two nodes is only affected by their sensitive attributes (\textbf{Assumption 2}). Here, the expression $N^{l \rightarrow k}N^{l \rightarrow l}(\tilde{E}^{(l,k)}-E^{(l,l)})$ represents the contribution from pairs of nodes where one kept $l$ as sensitive attribute and the other switched from $l$ to $k$.

Finally, we consider the case where both nodes switch their attributes. Let $S_i=S_j=l$ be the original attribute, and $Z_i=k_1$, $Z_j=k_2$ where $k_1 \ne l$ and $k_2 \ne l$. This corresponds to two nodes originally in class $l$ switching to classes $k_1$ and $k_2$:
\begin{align}
\begin{split}
&\sum_{i=1}^{N}\sum_{j=1}^{N}(\tilde{Y}_{ij}-Y_{ij})\mathbb{I}_{\{S_{i}=S_{j},Z_{j}\ne S_{j},Z_{i}\ne S_{i}\}} \\
    &= \sum_{l=1}^{K}\sum_{k_{1}=1; k_{1} \neq l}^{K} \sum_{k_{2}=1; k_{2}\neq l}^{K}\sum_{i,j}(\tilde{Y}_{ij}-Y_{ij})\mathbb{I}_{\{S_{i}=l, S_{j}=l, Z_{i}=k_{1}, Z_{j}=k_{2}\}} \\
    &\approx \sum_{l=1}^{K}\sum_{k_{1}=1; k_{1} \neq l}^{K} \sum_{k_{2}=1; k_{2}\neq l}^{K} N^{l \rightarrow k_{1}}N^{l \rightarrow k_{2}}(\tilde{E}^{(k_{1},k_{2})}-E^{(l,l)}).
\end{split}
\end{align}

We can eventually calculate the expectation over $\Delta_{\text{int}}$:
\begin{align}
\begin{split}
\mathbb{E}[\Delta_{\text{int}}] &= \mathbb{E}\Bigg[2\sum_{l=1}^{K}\sum_{k=1, k\ne l}^{K} N^{l \rightarrow k}N^{l \rightarrow l}(\tilde{E}^{(l,k)}-E^{(l,l)}) + \\
& \quad + \sum_{l=1}^{K}\sum_{k_{1}=1; k_{1} \neq l}^{K} \sum_{k_{2}=1; k_{2}\neq l}^{K} N^{l \rightarrow k_{1}}N^{l \rightarrow k_{2}}(\tilde{E}^{(k_{1},k_{2})}-E^{(l,l)})\Bigg] \\
& \text{(all random variables are independent)}\\
    &= 2\sum_{l=1}^{K}\sum_{k=1, k\ne l}^{K} \mathbb{E}[N^{l \rightarrow k}]\mathbb{E}[N^{l \rightarrow l}](\mathbb{E}[\tilde{E}^{(l,k)}]-\mathbb{E}[E^{(l,l)}]) + \\
     & \quad + \sum_{l=1}^{K}\sum_{k_{1}=1; k_{1} \neq l}^{K} \sum_{k_{2}=1; k_{2}\neq l}^{K} \mathbb{E}[N^{l \rightarrow k_{1}}]\mathbb{E}[N^{l \rightarrow k_{2}}](\mathbb{E}[\tilde{E}^{(k_{1},k_{2})}]-\mathbb{E}[E^{(l,l)}]),
\end{split}
\end{align}
where the first expectation term $\mathbb{E}\left[\sum_{i=1}^{N}\sum_{j=1}^{N}(Y^*_{ij}-Y_{ij})\mathbb{I}_{\{S_{i}=S_{j}=Z_{i}=Z_{j}\}}\right]$ was omitted as in (\Cref{eq:first_term}) we assumed the expectation is equal to zero. In the equation above, the expected number of nodes switching from attribute $l$ to $k$ ($k \ne l$) is given by:
\begin{equation}
\mathbb{E}[N^{l \rightarrow k}] = \rho \;  p_{\sigma}(k|l) \; N^{l},
\end{equation}
where $\rho$ represents, for each node, the probability of being switched (a Bernoulli variable parameter), $p_{\sigma}(k|l)$ is the conditional probability of sampling/assigning attribute $k \sim \mathcal{D}_\sigma(\mathcal{S} \setminus \{l\})$ given the original attribute $l$ and derived from the switching method used (i.e., Uniform or Prior), while $N^{l}$ is the total number of nodes having original sensitive attribute $l$. Conversely, the expected number of nodes originally with attribute $l$ that do not switch is:
\begin{equation}
\mathbb{E}[N^{l \rightarrow l}] = (1-\rho) N^{l}.
\end{equation}

The expectation of having a link between different sensitive attribute classes ($\mathbb{E}[E^{(l,k)}]$) can be estimated from the prior distribution of edges in the original graph based on its sensitive attributes. Substituting the expected values of $N^{l \rightarrow k}$ and $N^{l \rightarrow l}$ into the expression for $\mathbb{E}[\Delta_{\text{int}}]$:
\begin{align}
\begin{split}
\label{eq:delta_int}
\mathbb{E}[\Delta_{\text{int}}] &= 2\sum_{l=1}^{K}\sum_{k=1, k\ne l}^{K} \rho \; p_{\sigma}(k | l) N^{l} (1-\rho) N^{l} (\mathbb{E}[\tilde{E}^{(l,k)}]-\mathbb{E}[E^{(l,l)}]) + \\
    & \quad + \sum_{l=1}^{K}\sum_{k_{1}=1; k_{1} \neq l}^{K} \sum_{k_{2}=1; k_{2}\neq l}^{K} (\rho \; p_{\sigma}(k_{1}|l) N^{l}) (\rho \; p_{\sigma}(k_{2}|l) N^{l}) (\mathbb{E}[\tilde{E}^{(k_{1},k_{2})}]-\mathbb{E}[E^{(l,l)}]) \\
    &= 2\rho(1-\rho)\sum_{l=1}^{K}\sum_{k=1, k\ne l}^{K} p_{\sigma}(k|l)(N^{l})^{2}(\mathbb{E}[\tilde{E}^{(l,k)}]-\mathbb{E}[E^{(l,l)}])+ \\
    & \quad + \rho^2 \sum_{l=1}^{K}\sum_{k_{1}=1; k_{1} \neq l}^{K} \sum_{k_{2}=1; k_{2}\neq l}^{K} p_{\sigma}(k_{1}|l)p_{\sigma}(k_{2}|l)(N^{l})^2(\mathbb{E}[\tilde{E}^{(k_{1},k_{2})}]-\mathbb{E}[E^{(l,l)}]).
\end{split}
\end{align}

Finally, the expected number of new interior edges generated with switching is obtained by summing the number of original interior links and their expected difference:
\begin{equation}
\label{eq:int}
\mathbb{E}[|\tilde{\mathcal{E}}_{\text{int}}|] = \underbrace{\mathbb{E}\left[\sum_{i=1}^{N}\sum_{j=1}^{N}Y_{ij}\mathbb{I}_{\{S_{i}=S_{j}\}}\right]}_{= |\mathcal{E}_{\text{int}}| \text{ for \textbf{Assumption 1}}} + \mathbb{E}[\Delta_{\text{int}}] = |\mathcal{E}_{\text{int}}| + \mathbb{E}[\Delta_{\text{int}}].
\end{equation}

After some re-working of \Cref{eq:delta_int} and \Cref{eq:int} by grouping same-order coefficients together, we obtain a second-order equation in $\rho$:
\begin{equation}
    \mathbb{E}[|\tilde{\mathcal{E}}_{\text{int}}|] = R^2_{\text{int}}\rho^2 + R^1_{\text{int}}\rho + |\mathcal{E}_{\text{int}}|,
\end{equation}
where $R^2_{\text{int}}$ and $R^1_{\text{int}}$ are the short form representations for the coefficients of second and first order of the equation. 

\textbf{Exterior links.} Dually, let us consider the number of exterior edges after switching, $|\tilde{\mathcal{E}}_{\text{ext}}|$, where connected nodes have different real sensitive attributes $S_i \neq S_j$:
\begin{equation}
|\tilde{\mathcal{E}}_{\text{ext}}| =  \sum_{i=1}^{N}\sum_{j=1}^{N}Y_{ij}\mathbb{I}_{\{S_{i} \neq S_{j}\}} + \sum_{i=1}^{N}\sum_{j=1}^{N}(\tilde{Y}_{ij}-Y_{ij})\mathbb{I}_{\{S_{i} \neq S_{j}\}}.
\end{equation}
Once again, let $\Delta_{\text{ext}}$ be the difference term for exterior edges:
\begin{equation}
\Delta_{\text{ext}} = \sum_{i=1}^{N}\sum_{j=1}^{N}(\tilde{Y}_{ij}-Y_{ij})\mathbb{I}_{\{S_{i} \neq S_{j}\}}.
\end{equation}
Using the same logic of decomposing the indicator based on switching outcomes ($Z_i, Z_j$ vs. $S_i, S_j$) for the $S_i \neq S_j$ case, we end up with an expected value $\mathbb{E}[\Delta_{\text{ext}}]$:
\begin{align}
\begin{split}
    \mathbb{E}[\Delta_{\text{ext}}] &=  2\rho(1-\rho)\sum_{l_{1}=1}^{K}\sum_{l_{2}=1, l_{2}\neq l_{1}}^{K}\sum_{k=1, k\ne l_1}^{K} p_\sigma(k|l_1)N^{l_{1}}N^{l_{2}}(\mathbb{E}[\tilde{E}^{(k,l_{2})}]-\mathbb{E}[E^{(l_{1},l_{2})}])+ \\ 
    &\quad+ \rho^2 \sum_{l_{1}=1}^{K}\sum_{l_{2}=1, l_{2}\neq l_{1}}^{K}\sum_{k_{1}=1; k_{1} \neq l_{1}}^{K} \sum_{k_{2}=1; k_{2}\neq l_{2}}^{K} p_ \sigma(k_{1}|l_{1})p_\sigma(k_{2}|l_{2}) N^{l_{1}}N^{l_{2}} (\mathbb{E}[\tilde{E}^{(k_{1},k_{2})}]-\mathbb{E}[E^{(l_{1},l_{2})}]).
\end{split}
\end{align}
Finally, and as done in the case of interior links, we add the expected number of original exterior edges to the expectation of the difference, and group same-order coefficients such as:
\begin{equation}
    \mathbb{E}[|\tilde{\mathcal{E}}_{\text{ext}}|] = R^2_{\text{ext}}\rho^2 + R^1_{\text{ext}}\rho + |\mathcal{E}_{\text{ext}}|.
\end{equation}

\textbf{Overall optimization problem.} In our intuition, we can reach fairness by balancing the difference of the expected exterior and interior links generated with switching and their difference in the original graph. Concretely, if the difference between exterior and interior links in the original graph is positive (negative), we aim to find $\rho^*$ such that the same difference in the generated graph with switching will go in the opposite direction, being minimized (maximized). In other terms, switching on a fraction of nodes $\rho^*$ will balance the bias amplification from the GDM. Our optimization problem becomes:
\begin{equation}
\rho^{*} = \argmin_{\rho \in [0, 1]} \text{sign}(R_0)(R_2\rho^{2} + R_1\rho + R_0),
\end{equation}
where $R_2 =  R_{\text{ext}}^{2} - R_{\text{int}}^{2}$, $R_1 = R_{\text{ext}}^{1} - R_{\text{int}}^{1}$, $R_0 = |\mathcal{E}_{\text{ext}}| - |\mathcal{E}_{\text{int}}|$, while $\text{sign}(R_0)$ stands for the sign of $R_0$ and serves to drive the optimization in the opposite direction of the imbalance.

\subsection{Fused Gromov-Wasserstein Distance}

In this section, we provide a more complete explanation of the Fused Gromov-Wasserstein distance (FGW), and how we adapted it to assess the differences between the original and any generated graph in the FAROS pipeline.

The FGW distance provides a way to compare structured objects, such as graphs, by simultaneously considering both their node features and their underlying structure. Unlike the standard Wasserstein distance~\cite{Wasserstein}, which focuses solely on feature similarity, or the Gromov-Wasserstein distance~\cite{DBLP:journals/focm/Memoli11, DBLP:journals/tog/SolomonPKS16}, which only compares structural relationships, FGW integrates both aspects.

Let us consider two graphs $\mathcal{G}_{\mu} = \{\mathcal{V}_{\mu}, \mathcal{E}_{\mu}\}$ and $\mathcal{G}_{\nu} = \{\mathcal{V}_{\nu}, \mathcal{E}_{\nu}\}$, where $|\mathcal{V}_{\mu}| = N$ and $|\mathcal{V}_{\nu}| = M$, respectively. In this context, each graph is represented by its probability measures  $\mu=\sum_{n=1}^{N}h_{n}\delta_{(x_{n},a_{n})}$ and $\nu=\sum_{m}^{M}g_{m}\delta_{(y_{m},b_{m})}$, where $(x_{n}, a_{n})$ and $(y_{m}, b_{m})$ represent the feature and structure components of nodes $n$ and $m$ with histograms $h_{n} \in \sum_n$ and $g_{m} \in \sum_m$. Then, the Wasserstein distance between the two graphs adopts a transport cost matrix $\mathbf{M} \in \mathbb{R}^{N \times M}$ between locations in $\mathcal{G}_{h}$ and $\mathcal{G}_{g}$ (considering only the feature components), while the Gromov-Wasserstein distance adopts internal transportation cost matrices $\mathbf{C}_{h} \in \mathbb{R}^{N \times N}$ and $\mathbf{C}_{g} \in \mathbb{R}^{M \times M}$ for transportation within the same graph (considering only the structure component). 

On such basis, the FGW distance seeks an optimal probabilistic matching (coupling) $\mathbf{P} \in \Pi(h,g)$ between the nodes of the two graphs that minimizes the overall cost function:
\begin{equation}
    \text{FGW}(h, g) = \min_{\mathbf{P} \in \prod(h, g)} \langle \alpha\mathbf{M}+(1-\alpha)|\mathbf{C}_{h}(i, k)-\mathbf{C}_{g}(j, l)|^2 \otimes \mathbf{P}, \mathbf{P} \rangle,
\end{equation}
where we indicate as the set of all admissible couplings between $h$ and $g$:
\begin{equation}
\Pi(h, g) = \left\{\mathbf{P} \in \mathbb{R}^{N \times M} \text{ s.t. } \sum_{n=1}^{N} P_{n,m} = h_m, \sum_{m=1}^{M} P_{n,m} = g_{n} \right\}.
\end{equation}

This cost is a linear combination, controlled by a trade-off parameter $\alpha \in [0,1]$, balancing the Wasserstein distance and the Gromov-Wasserstein distance (obtained for $\alpha = 1$ and $\alpha = 0$, respectively).

In our setting, we calculate the FGW distance between the original graph and any generated graph through the GDM, so $\mathcal{G}_h := \mathcal{G}$ and $\mathcal{G}_g := \tilde{\mathcal{G}}$. Moreover, it is important to highlight that, by design~\cite{DBLP:journals/corr/abs-2310-13833}, the two graphs have the same number of nodes, so the above cited matrices are re-defined as $\mathbf{M} \in \mathbb{R}^{N \times N}$, $\mathbf{C}_{h}, \mathbf{C}_{g}$, $\mathbf{P} \in \mathbb{R}^{N \times N}$. Note also that, in the main paper (refer again to \Cref{sec:multi-criteria}), we followed a slight notation abuse by replacing $h$ and $g$ with the corresponding graphs $\mathcal{G}$ and $\tilde{\mathcal{G}}$. This was done to ease the notation and in the interest of space. 
\section{Detailed Experimental Settings}
\label{sec:add_settings}

\subsection{Datasets}

\Cref{tab:datasets} reports on the statistics of the three graph datasets we used for this work. As our code base was built on top of FairWire\footnote{Accessible on OpenReview under the Supplementary Material link: \url{https://tinyurl.com/vrp6n9w7}.}~\cite{DBLP:conf/nips/KoseS24}, we decided to adopt their datasets for the sake of fair comparison to their approach. For this reason, the processing of datasets is the same as the one followed in their paper. Specifically, for \textsc{Cora} and \textsc{Citeseer} (two citation networks~\cite{DBLP:journals/aim/SenNBGGE08}) we adopted one-hot encoding of the original paper descriptions as binary node features, where the category of the nodes (papers) are used as the sensitive attribute. For the third dataset \textsc{Amazon Photo} (a co-purchase network~\cite{DBLP:journals/corr/abs-1811-05868}) we set again one-hot encoding of the product reviews as the node features, and the product categories as sensitive attribute. 

\begin{table}[!h]
\caption{Datasets statistics.}\label{tab:datasets} 
\centering 
\footnotesize
\begin{tabular}{lrrrr}
\toprule
\textbf{Datasets} & $|\mathcal{V}|$ & $|\mathcal{E}|$ & $F$ & $K$ \\
\cmidrule{1-5} 
\textsc{Cora} & 2,708 & 10,556 & 1,433 & 7 \\
\textsc{Citeseer} & 3,327 & 9,228 & 3,703 & 6 \\
\textsc{Amazon Photo} & 7,650 & 238,163 & 745 & 8 \\
\bottomrule
\end{tabular}
\end{table}

\subsection{Baselines}

As baselines, and following~\cite{DBLP:conf/nips/KoseS24}, we selected GraphMaker~\cite{DBLP:journals/corr/abs-2310-13833} (in the synchronous version), FairAdj~\cite{DBLP:conf/iclr/LiWZHL21}, the adversarial regularizer proposed in~\cite{DBLP:conf/wsdm/DaiW21} (denoted as Adversarial in our work), and FairWire~\cite{DBLP:conf/nips/KoseS24}. We re-trained from scratch GraphMaker and  FairWire using the same codes and settings as reported in the original codebase of FairWire. Conversely, for FairAdj and Adversarial, we directly reported the obtained results from the FairWire paper~\cite{DBLP:conf/nips/KoseS24}, as the available codebase does not include their implementations within the graph generation/link prediction pipeline. Even assuming that a complete re-training of all baselines would have not lead to drastically-different trends with respect to those reported in the original paper~\cite{DBLP:conf/nips/KoseS24}, we decided to re-train from scratch FairWire anyway to ensure its results were fully comparable to those we computed of FAROS, as it represents the most similar and competitive baseline to our approach. Finally, regarding FAROS, we implemented and tested two variants with Uniform and Prior sampling distributions (referred to as FAROS-Uniform and FAROS-Prior, respectively). 

In the specific setting of RQ3 (last part of \Cref{sec:results}), we applied FAROS to GraphMaker synchronous version (GraphMaker-Sync, the same setting described above), asynchronous version (GraphMaker-Async), and FairWire. For GraphMaker-Async, we mounted FAROS on the pre-trained asynchronous version of GraphMaker, which generates node features and graph edges in alternating time steps; thus, we apply the attribute switching procedure just in the time steps for the edges generation. Finally, in the case of FairWire, we used it along the pre-trained GraphMaker-Sync diffusion model leveraging the fairness regularizer proposed in FairWire.

\subsection{Hyper-parameters} 

One of the advantages of our approach, FAROS, is that it does not require any re-training of the GDM it is built on. For this reason, we did not need to perform large hyper-parameter explorations, apart from searching the optimal switching time $\tau^*$, which was selected through grid-search in the range $\{1, 2, 3\}$ to optimize the objective function in \Cref{eq:opt_tau}. We highlight that this set of values comes from GraphMaker's experimental settings (in the synchronous version), where the authors adopt $T = 3$ as total number of diffusion steps. Thus, when selecting $\tau^* = 3$, it corresponds to not performing any attribute switching, which is conceptually and empirically equal to running the original GraphMaker algorithm. Noteworthy, our grid-search steadily selected $\tau^* = 2$ for all datasets and sampling distributions (Uniform and Prior), but never $\tau^* = 3$. This further demonstrates how FAROS can be effectively mounted on GraphMaker and improve the accuracy-fairness trade-off. Finally, within \Cref{eq:opt_tau}, we also selected the coefficient $\gamma$ (balancing the accuracy and fairness contributions) within $\{0.0, 0.1, \ldots, 1.0\}$, empirically finding a good balance with $\gamma = 0.5$.

To exploit FAROS in combination with GraphMaker-Async (RQ3), we followed slightly different settings from those reported above. Specifically, in this case, we searched $\tau^*$ within the set $\{1, 2, \ldots, 9\}$, where again $\tau^* = 9$ stands for no switching (thus, the original algorithm proposed in GraphMaker-Async). Once again, this set of values was selected based upon the original paper of GraphMaker, as the authors propose to adopt $T_\mathbf{X} = 6$ and $T_{\mathcal{E}} = 9$, where the former and the latter stand for the number of time steps for the generation of nodes features and edges, respectively. In this case, our optimizaton method selected $\tau^* = 8$ for both Uniform and Prior on \textsc{Cora}, while $\tau^* = 7$ and $\tau^* = 6$ for Uniform and Prior on \textsc{Amazon Photo}. Similarly to the synchronous setting, we empirically found $\gamma = 0.5$ as the best option. 

\subsection{Training and evaluation pipelines} 

In the following, we enumerate all the steps to reproduce our complete pipeline and get the final performance results. To begin with, we re-trained from scratch all GDMs, namely, GraphMaker-Sync, GraphMaker-Async, and FairWire. Secondly, we used their checkpoints to generate 10 samples from each dataset, following the same experimental settings as in FairWire~\cite{DBLP:conf/nips/KoseS24}. On such a basis, we trained a link prediction model on each of these generated graphs, implemented as a 1-layer Graph Autoencoder (GAE). Finally, we tested the same model on the original test set from each dataset. In terms of evaluation metrics, we used the Area Under Curve (AUC) to assess prediction accuracy, and the Statistical Parity ($\Delta_{SP}$) and Equality of Opportunity ($\Delta_{EO}$) for the fairness metrics, where we seek to maximize and minimize the former and the latter, respectively.

\subsection{Computational resources}

All our experiments were run on a single Linux machine equipped with one AMD EPYC 7502 32-Core CPU, a RAM with 500 GB, and one Tesla V100S-PCIE GPU with 32 GB.  
\section{Additional Experimental Results}
\label{sec:add_experiments}

\subsection{Pareto optimality visualization}

\begin{figure}[!h]
    \centering
    \includegraphics[width=0.95\linewidth]{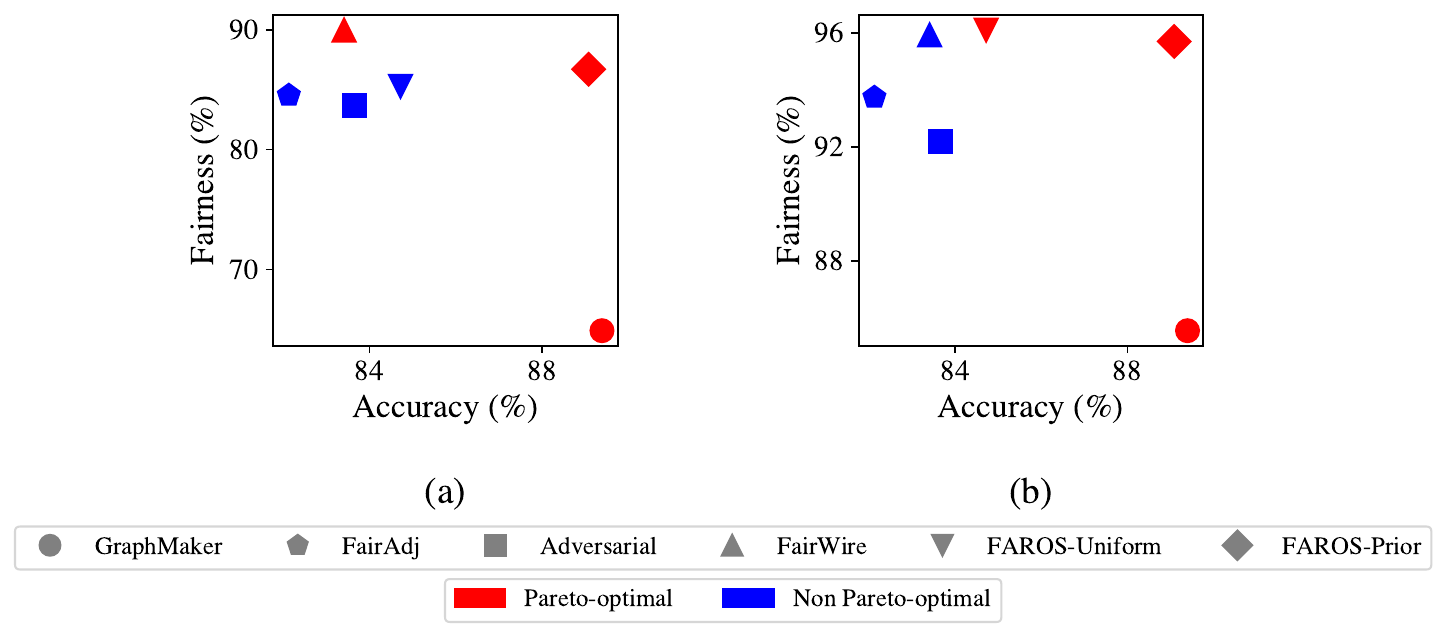}
    \caption{Pareto visualization on \textsc{Cora}, with (a) AUC vs. $\Delta_{SP}$ and (b) AUC vs. $\Delta_{EO}$.}
    \label{fig:pareto_cora}
\end{figure}

\begin{figure}[!h]
    \centering
    \includegraphics[width=0.95\linewidth]{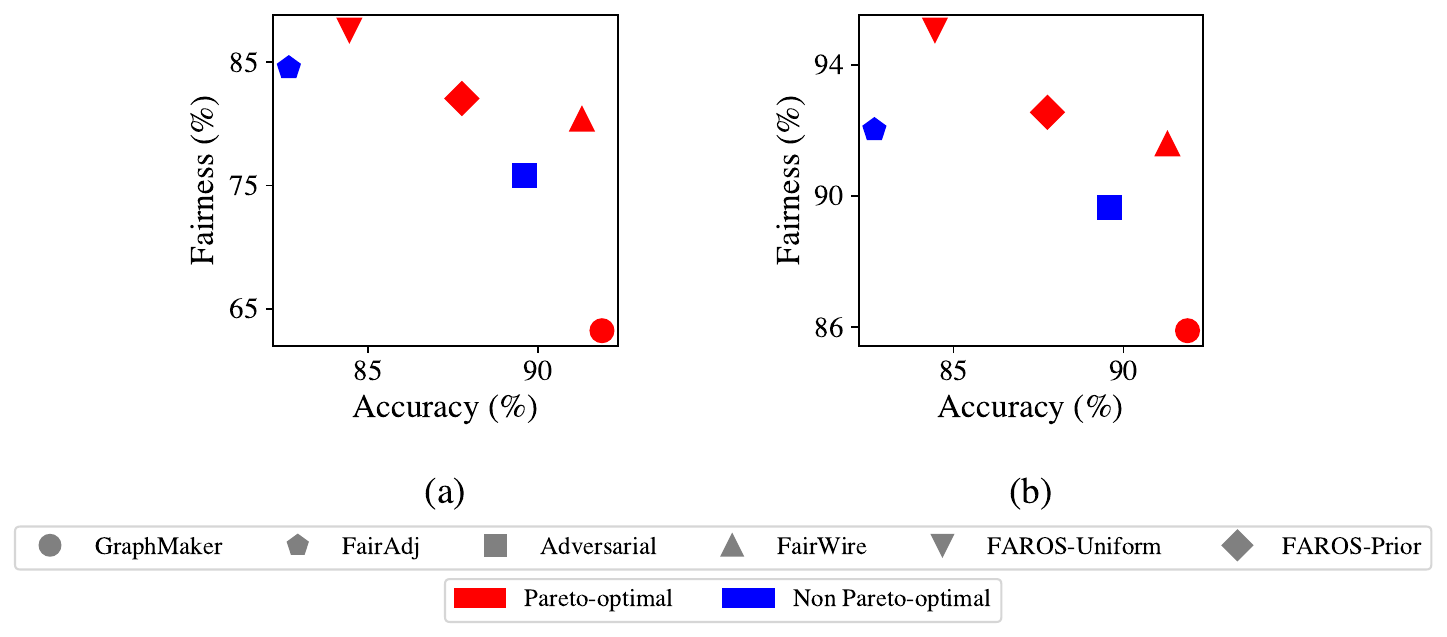}
    \caption{Pareto visualization on \textsc{Citeseer}, with (a) AUC vs. $\Delta_{SP}$ and (b) AUC vs. $\Delta_{EO}$.}
    \label{fig:pareto_citeseer}
\end{figure}

\begin{figure}[!h]
    \centering
    \includegraphics[width=0.8\linewidth]{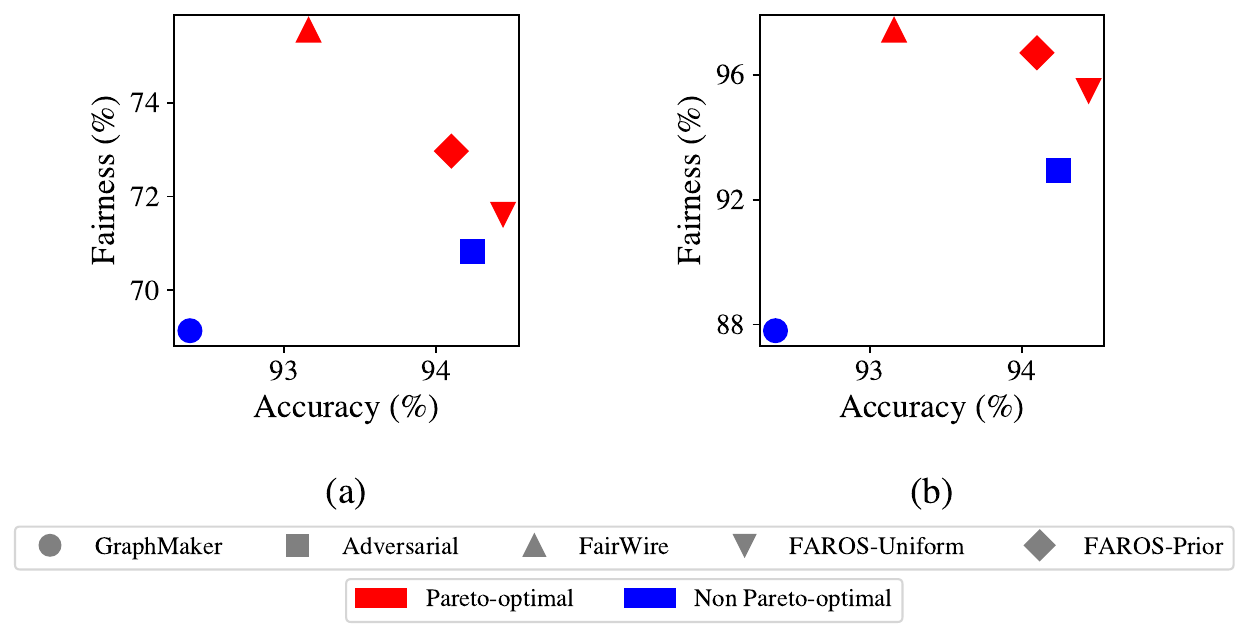}
    \caption{Pareto visualization on \textsc{Amazon Photo}, with (a) AUC vs. $\Delta_{SP}$ and (b) AUC vs. $\Delta_{EO}$. Results for FairAdj was not computed on \textsc{Amazon Photo}.}
    \label{fig:pareto_amazon-photo}
\end{figure}

In this section, we present all performance results of the tested baselines under the Pareto optimality definition. For all displayed plots (Figures \ref{fig:pareto_cora}-\ref{fig:pareto_citeseer}-\ref{fig:pareto_amazon-photo}), we compare accuracy (AUC) against fairness ($\Delta_{SP}$ and $\Delta_{EO}$), by explicitly highlighting in \textcolor{red}{\textbf{red}} those models which lay within the Pareto frontier and in \textcolor{blue}{\textbf{blue}} all other baselines. We believe that the proposed visualization could better underscore the performance trade-off between accuracy and fairness of the tested baselines, showing how our proposed approaches (FAROS-Uniform and FAROS-Prior) are in most cases able to strike a better trade-off than most of the other solutions.

\clearpage
\clearpage

For the sake of completeness, we briefly recall the concept of Pareto optimal solutions and Pareto frontier in multi-objective optimization. Let $f_i = \{f_1(\cdot), f_2(\cdot), \ldots, f_m(\cdot)\}$ be the set of all objective functions. Then, let $x \in \mathcal{X}$ be the set of all admissible solutions. On such a basis, in a multi-objective optimization setting where all objectives are maximized, we say $x_1$ dominates over $x_2$ (represented as $x_1 \succ x_2$) iff $f_i(x_1) \geq f_i(x_2) \; \forall f_i(\cdot)$ and $\exists f_j(\cdot) \text{ s.t. }f_j(x_1) > f_j(x_2)$. Thus, all $x$ which are not dominated by any other solutions will belong to the Pareto frontier.

In our setting, we have that $m = 2$, as we consider accuracy and fairness metrics in pairs (e.g., AUC vs. $\Delta_{SP}$). Moreover, since accuracy metrics and fairness metrics are supposed to be maximized and minimized, respectively, we rescale fairness metrics so that they also adhere to the principle ``the higher the better''. To this end, and only for this visualization, we rescale both fairness metrics by calculating $100 - \Delta_{SP}$ and $100 - \Delta_{EO}$. 

\subsection{Calculated optimal $\rho^*$ values}

In \Cref{tab:best_rho}, we show the calculated values (obtained through \Cref{eq:opt_frac}) of the node fraction to be switched during the generation of the GDM for all the tested datasets. As observable, percentages are lower than $100 \%$ in all settings, especially for FAROS-Prior.

\begin{table}[!h]
\caption{Calculated optimal fraction of nodes to be switched $\rho^*$ on all the tested datasets.}\label{tab:best_rho} 
\centering 
\begin{adjustbox}{max width=0.5\linewidth}
\begin{tabular}{lccc}
\toprule
\textbf{Models} & \textsc{Cora} & \textsc{Citeseer} & \textsc{Amazon Photo} \\ \cmidrule{1-4}
FAROS-Uniform & 90.02\% & 83.34\% & 97.67\% \\
FAROS-Prior & 81.56\% & 81.85\% & 83.14\% \\
\bottomrule
\end{tabular}
\end{adjustbox}
\end{table}

\subsection{Empirical justification of the optimal $\tau^*$ selection} 

\begin{figure}[!h]
    \centering    \includegraphics[width=0.6\columnwidth]{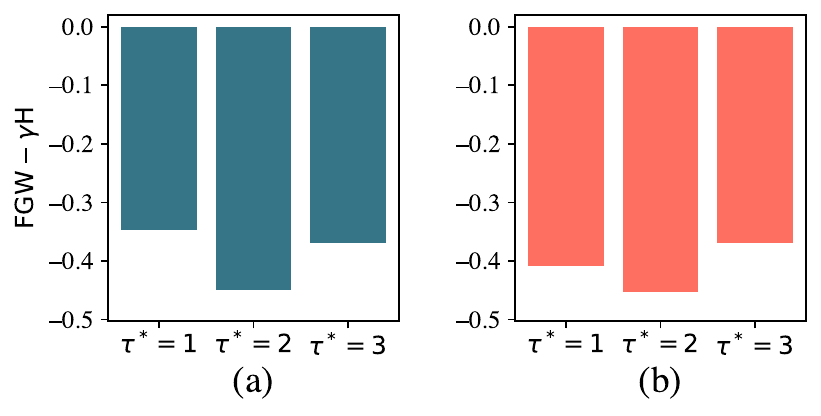} 
    \caption{Variation of the objective function (for $\gamma = 0.5$) on \textsc{Cora}, where (a) and (b) indicate results for FAROS-Uniform and FAROS-Prior, respectively.}
    \label{fig:loss_cora}
\end{figure}

\begin{figure}[!h]
    \centering    \includegraphics[width=0.6\columnwidth]{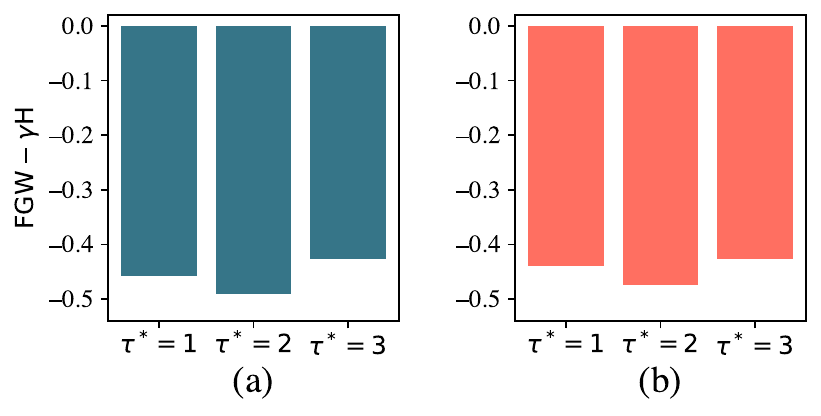} 
    \caption{Variation of the objective function (for $\gamma = 0.5$) on \textsc{Citeseer}, where (a) and (b) indicate results for FAROS-Uniform and FAROS-Prior, respectively.}
    \label{fig:loss_citeseer}
\end{figure}

According to our multi-criteria optimization method (refer again to \Cref{sec:multi-criteria}), the optimal time step to perform the switching mechanisms was selected as $\tau^*=2$ on all tested datasets. To empirically validate this outcome, we compute all link prediction results when manually setting $\tau^* = \{1, 2, 3\}$, our complete search space. Empirical results (\Cref{tab:tau_selection}) demonstrate that the settings for $\tau^* = 1, 3$ cannot reach a similar accuracy-fairness trade-off as in the case of $\tau^* = 2$, with both FAROS-Uniform and FAROS-Prior on \textsc{Cora} and \textsc{Citeseer}. Additionally, in Figures \ref{fig:loss_cora}-\ref{fig:loss_citeseer}, we plot the values of the multi-criteria objective function (\Cref{eq:opt_tau}) for both considered datasets. Once again, we notice how the lowest function value is reached at $\tau^* = 2$. All these outcomes further justify the goodness of our multi-criteria selection of $\tau^*$.

\begin{table}[!h]
\caption{Link prediction results of FAROS on \textsc{Cora} and \textsc{Citeseer} with different value of the time step $\tau^*$ when to perform the attribute switching.}\label{tab:tau_selection} 
\centering 
\begin{adjustbox}{max width=\columnwidth}
\begin{tabular}{llcccccc}
\toprule
\textbf{Distributions} &    \makecell[l]{\textbf{Optimal}\\\textbf{switching}\\\textbf{time}} & \multicolumn{3}{c}{\textsc{Cora}} & \multicolumn{3}{c}{\textsc{Citeseer}} \\ \cmidrule(lr){3-5} \cmidrule(lr){6-8}
& & AUC ($\uparrow$) & $\Delta_{SP}$ ($\downarrow$) & $\Delta_{EO}$ ($\downarrow$) & AUC ($\uparrow$) & $\Delta_{SP}$ ($\downarrow$) & $\Delta_{EO}$ ($\downarrow$) \\
\cmidrule{1-8}
\multirow{3}{*}{FAROS-Uniform} 
& $\tau^* = 1$ & 86.57$\pm$2.29 & 28.94$\pm$1.38 & 11.48$\pm$1.03 & 89.99$\pm$1.10 & 30.06$\pm$4.11 & 12.14$\pm$1.24 \\
& $\tau^* = 2$ & 84.72$\pm$5.65 & \textbf{14.73$\pm$6.62} & \textbf{3.90$\pm$2.81} & 84.45$\pm$7.71 & \textbf{12.39$\pm$8.05} & \textbf{4.94$\pm$3.30} \\
& $\tau^* = 3^\dagger$ & \textbf{89.39$\pm$0.92} & 35.12$\pm$2.19 & 14.45$\pm$0.77 & \textbf{91.88$\pm$0.98} & 36.77$\pm$1.22 & 14.11$\pm$0.94 \\
\cmidrule{1-8}
\multirow{3}{*}{FAROS-Prior} 
& $\tau^* = 1$ & \textbf{89.48$\pm$1.19} & 32.70$\pm$3.08 & 12.24$\pm$1.95 & 92.51$\pm$1.26 & 34.70$\pm$2.57 & 13.50$\pm$0.79 \\
& $\tau^* = 2$ & 89.08$\pm$2.72 & \textbf{13.30$\pm$8.62} & \textbf{4.30$\pm$4.03} & 87.76$\pm$6.21 & \textbf{17.94$\pm$9.46} & \textbf{7.45$\pm$3.65} \\
& $\tau^* = 3^\dagger$ & 89.39$\pm$0.92 & 35.12$\pm$2.19 & 14.45$\pm$0.77 & \textbf{91.88$\pm$0.98} & 36.77$\pm$1.22 & 14.11$\pm$0.94 \\
\bottomrule
\multicolumn{8}{l}{\footnotesize$^\dagger$ Corresponding to the no switching setting, the same as GraphMaker.}
\end{tabular}
\end{adjustbox}
\end{table}

\subsection{Testing other diffusion models on \textsc{Amazon Photo}}

We extend the experimental settings proposed in RQ3 (last part of \Cref{sec:results}) by running further experiments on \textsc{Amazon Photo}, when selecting three GDMs: GraphMaker-Sync, GraphMaker-Async, and FairWire. Results in \Cref{tab:diffusion_backbones_2} generally confirm what we already observed in RQ3. That is, the adoption of GraphMaker-Async provides a quite strong baseline reaching a better trade-off than GraphMaker-Sync and FairWire. Notably, and unlike the discussed outcomes in RQ3 for \textsc{Cora}, here we notice that the adoption of FAROS-Prior on FairWire can provide improved link prediction performance than leveraging FairWire alone. As a matter of facts, in this case, our multi-criteria optimization method suggested to perform attribute switching, correctly leading to better link prediction results (in \Cref{tab:diffusion_backbones_2}, compare this trade-off with that reached by FairWire and FAROS-Uniform, where no switched is performed). The observed trends highlight how, in specific settings, FAROS could effectively support even fairness-aware generators. This further motivates us to explore the effects of FAROS on other GDMs, both fairness-agnostic and fairness-aware ones. 

\begin{table}[!h]
\caption{Link prediction results for FAROS on \textsc{Amazon Photo} with different diffusion backbones. }\label{tab:diffusion_backbones_2} 
\centering 
\footnotesize
\begin{tabular}{llccc}
\toprule
\textbf{Distributions} & \makecell[l]{\textbf{Diffusion}\\\textbf{models}} & AUC ($\uparrow$) & $\Delta_{SP}$ ($\downarrow$) & $\Delta_{EO}$ ($\downarrow$) \\
\cmidrule{1-5} 
\multirow{3}{*}{FAROS-Uniform} & GraphMaker-Sync & \textbf{94.44$\pm$0.28} & 28.38$\pm$1.52 & 4.50$\pm$0.78 \\
& GraphMaker-Async & 94.22$\pm$0.34 & \textbf{20.37$\pm$7.93} & 6.06$\pm$2.25 \\
& FairWire & 93.16$\pm$0.32 & 24.44$\pm$1.06 & \textbf{2.56$\pm$0.4}7 \\
\cmidrule{1-5}
\multirow{3}{*}{FAROS-Prior} & GraphMaker-Sync & 94.10$\pm$0.25 & 27.03$\pm$0.36 & 3.29$\pm$0.27 \\
& GraphMaker-Async & \textbf{94.50$\pm$0.32} & 23.55$\pm$5.94 & 7.15$\pm$2.05 \\
& FairWire$^\dagger$ & 92.26$\pm$1.33 & \textbf{23.11$\pm$3.37} & \textbf{1.79$\pm$0.89} \\
\bottomrule
\multicolumn{5}{l}{\footnotesize$^\dagger$ In this setting, FairWire has been effectively supported by FAROS.}
\end{tabular}
\end{table}

\subsection{Topological differences}

To conclude, we provide results on the generation capabilities of GraphMaker-Sync when combined with FAROS-Uniform and FAROS-Prior. \Cref{tab:topology} outlines topological differences between the original and the generated graphs in terms of 1-Wasserstein difference of the two probability distributions, for the degree (Deg.) and the clustering coefficient (Clust.). Noteworthy, we observe that both variants of FAROS can lead to generated graphs whose topology is close to the original one, while still ensuring fairness. This trend keeps steady on all datasets, where FAROS is always best or second-to-best and outperforms FairWire, in some cases even by a large margin (e.g., \textsc{Amazon Photo}). These results empirically demonstrate the ability of FAROS to precisely intervene within the generation process to drive it towards fairness, and affecting it the least. We maintain such good results with FAROS could also be ascribed to the multi-criteria optimization method for the selection of $\tau^*$, where the objective component accounting for the accuracy (i.e., FGW) considers both topology and node features. 

\begin{table}[!h]
\caption{Topological differences between the original and the generated graphs, calculated as the 1-Wasserstein difference between the two probability distributions.}\label{tab:topology} 
\centering 
\begin{tabular}{lcccccc}
\toprule
\textbf{Models} & \multicolumn{2}{c}{\textsc{Cora}} & \multicolumn{2}{c}{\textsc{Citeseer}} & \multicolumn{2}{c}{\textsc{Amazon Photo}} \\ \cmidrule(lr){2-3} \cmidrule(lr){4-5} \cmidrule(lr){6-7}
& Deg. ($\downarrow$) & Clust. ($\downarrow$) & Deg. ($\downarrow$) & Clust. ($\downarrow$) & Deg. ($\downarrow$) & Clust. ($\downarrow$) \\ \cmidrule{1-7}
GraphMaker & \textbf{23.05} & \underline{0.57} & \textbf{13.79} & \underline{0.68} & \textbf{1.34} & 113.39 \\
FairWire & 23.75 & 1.75 & 13.95 & 0.94 & 1.56 & 73.09 \\
\cmidrule{1-7}
FAROS-Uniform & \underline{23.55} & \textbf{0.56} & \underline{13.88} & \textbf{0.67} &  1.48 & \textbf{7.72} \\
FAROS-Prior & 23.59 & 0.94 & 13.91 & \underline{0.68} & \underline{1.47} & \underline{12.27} \\
\bottomrule
\end{tabular}
\end{table}

\section{Broader Impacts}
\label{sec:broader_impact}

In the paper, we have proposed FAROS, a novel approach to address fairness issues in graph diffusion models (GDMs) by driving the generation of the GDM towards unbiased generated graph data. We believe our contribution can provide broader impacts if we consider real-world societal applications of GDMs. Such approaches, especially when exploited for large graph generation, can serve as tools to generate new synthetic data from the original one by preserving privacy (e.g., a critical concern in social media platforms) and tackling noisy/missing information in the original data (e.g., a critical concern in the healthcare domain). However, as existing GDMs have been shown to lead to biased generated data, FAROS would counteract this negative effect, allowing the adoption of GDMs for the outlined scopes without incurring in additional risks.

\end{document}